\documentclass{article}
\usepackage{PRIMEarxiv}

\usepackage{times}
\usepackage{helvet}
\usepackage{courier}
\usepackage{graphicx}
\usepackage{caption}

\usepackage{algorithm}
\usepackage{algorithmic}
\usepackage{subcaption}
\usepackage{amsfonts}
\usepackage{amssymb}
\usepackage{soul}
\usepackage{amsmath}
\usepackage{amsthm}
\usepackage{mathtools}
\usepackage{placeins}
\usepackage{enumerate}
\usepackage{mathabx}
\usepackage{mathrsfs}
\usepackage{placeins}
\usepackage{stmaryrd}

\SetSymbolFont{stmry}{bold}{U}{stmry}{m}{n}

\theoremstyle{plain}
\newtheorem{theorem}{Theorem}
\newtheorem{proposition}{Proposition}
\newtheorem{lemma}{Lemma}

\theoremstyle{definition}
\newtheorem{definition}{Definition}
\newtheorem{assumption}{Assumption}

\theoremstyle{remark}
\newtheorem{remark}{Remark}

\DeclareMathOperator{\argmax}{arg\,max}
\DeclareMathOperator{\argmin}{arg\,min}
\DeclareMathOperator{\Lip}{Lip}
\DeclareMathOperator{\subjectto}{s.t.}

\title{Wasserstein Distributionally Robust Multiclass Support Vector Machine}

\author{
  Michael Ibrahim, Heraldo Rozas, Nagi Gebraeel\\
  H. Milton Stewart School of Industrial and Systems Engineering \\
  Georgia Institute of Technology \\
  Atlanta\\
  \texttt{\{mibrahim41,heraldo.rozas,nagi\}@gatech.edu} \\
}

\begin{document}
\maketitle

\vspace{-20pt}

\begin{abstract}
We study the problem of multiclass classification for settings where data features $\mathbf{x}$ and their labels $\mathbf{y}$ are uncertain. We identify that distributionally robust one-vs-all (OVA) classifiers often struggle in settings with imbalanced data. To address this issue, we use Wasserstein distributionally robust optimization to develop a robust version of the multiclass support vector machine (SVM) characterized by the Crammer-Singer (CS) loss. First, we prove that the CS loss is bounded from above by a Lipschitz continuous function for all $\mathbf{x} \in \mathcal{X}$ and $\mathbf{y} \in \mathcal{Y}$, then we exploit strong duality results to express the dual of the worst-case risk problem, and we show that the worst-case risk minimization problem admits a tractable convex reformulation due to the regularity of the CS loss. Moreover, we develop a kernel version of our proposed model to account for nonlinear class separation, and we show that it admits a tractable convex upper bound. We also propose a projected subgradient method algorithm for a special case of our proposed linear model to improve scalability. Our numerical experiments demonstrate that our model outperforms state-of-the art OVA models in settings where the training data is highly imbalanced. We also show through experiments on popular real-world datasets that our proposed model often outperforms its regularized counterpart as the first accounts for uncertain labels unlike the latter.
\end{abstract}

\section{Introduction} \label{sec:intro}
Multiclass classification models are widely used to inform decision-making in diverse application domains, including disease diagnosis \cite{Zhang2021,Altunbay2010,Zhu2014}, fault diagnostics in industrial settings \cite{Li2016,Wang2020}, and cybersecurity \cite{toupas2019,Liu2024,Hussein2023}. Classification models aim to predict a discrete label $\mathbf{y} \in \mathcal{Y}$ from input features $\mathbf{x} \in \mathcal{X} \subseteq \mathbb{R}^P$. Such models are trained in a supervised fashion using a loss function $\ell(\mathbf{M};(\mathbf{x},\mathbf{y}))$ parameterized by $\mathbf{M} \in \mathcal{M}$, which may be a vector or a matrix depending on the type of classifier. Different families of loss functions give rise to different classification models. In many applications, both the input data features $\mathbf{x}$ and their labels $\mathbf{y}$ represent random variables \cite{shafieezadehabadeh2015distributionally}. Input data features are often corrupted by noise \cite{García2015}, whereas their corresponding labels can be erroneously annotated \cite{adebayo2023quantifying,northcutt2021pervasive}. This uncertainty often hurts the out-of-sample performance of classical classifiers. 

In this work, we denote the random variable that represents the tuple of input features and its corresponding label as  $\boldsymbol{\xi}=(\mathbf{x},\mathbf{y})$. In training, we seek to obtain the optimal parameters $\mathbf{M}^{\ast}$ that minimize the expected risk $\mathbb{E}^{\mathbb{P}}[\ell(\mathbf{M};\boldsymbol{\xi})]$, where $\mathbb{P}$ is the true distribution governing $\boldsymbol{\xi}$. It can be easily seen that the expected risk minimization problem is an example of stochastic programming (SP), which is thoroughly studied in \cite{shapiro_stochprog}. However, $\mathbb{P}$ is often unknown in most real-world applications \cite{rui2023,kuhn2019wasserstein}. This renders the expected risk minimization problem impossible to solve in practice. 

A popular alternative to minimizing the expected risk in practical applications is Sample Average Approximation (SAA) \cite{kleywegt2002}. This method assumes that there is a set of $N$ i.i.d. samples denoted as $\{ \widehat{\boldsymbol{\xi}}_n \}_{n=1}^{N}$  drawn from $\mathbb{P}$. These samples are used to estimate $\widehat{\mathbb{P}}_N$. Subsequently, one can obtain an approximation $\mathbf{M}^{\ast}_{\text{emp}}$ of the optimal model parameters by minimizing the empirical risk $\mathbb{E}^{\widehat{\mathbb{P}}_N}[\ell(\mathbf{M};\boldsymbol{\xi})]$. The empirical distribution of $\{ \widehat{\boldsymbol{\xi}}_n \}_{n=1}^{N}$ or an elliptical distribution whose parameters are estimated via maximum likelihood estimation (MLE) using $\{ \widehat{\boldsymbol{\xi}}_n \}_{n=1}^{N}$ are popular choices of $\widehat{\mathbb{P}}_N$ in the discrete and continuous settings, respectively. It has been shown in \cite{kuhn2019wasserstein} that these choices are asymptotically optimal within their respective families in approximating $\mathbb{P}$, attaining the fastest convergence rate to $\mathbb{P}$ as $N \rightarrow \infty$. Nonetheless, $\widehat{\mathbb{P}}_N$ may still be a poor approximation of $\mathbb{P}$ for limited sample sizes $N$, thereby failing to characterize the actual uncertainty in $\boldsymbol{\xi}$. This will result in a model that suffers from the \textit{optimizer's curse} \cite{james2006,kuhn2019wasserstein,Li2021}--the model will achieve very low loss on the training set, but will exhibit poor out-of-sample performance.

The recently reemerging field of Distributionally robust optimization (DRO) \cite{scarf1957min,bental2013,bayraksan2015,shapiro2017,datadrivenDRO,rui2023} aims to improve out-of-sample performance by hedging against the uncertainty in $ \boldsymbol{\xi}$. This is done by constructing an ambiguity set $\mathcal{A}$ and obtaining an improved approximation $\mathbf{M}^{\ast}_{\text{dro}}$ of the optimal model parameters by minimizing the worst-case risk $\sup_{\mathbb{Q} \in \mathcal{A}} \mathbb{E}^{\mathbb{Q}}[\ell(\mathbf{M};\boldsymbol{\xi})]$. The ambiguity set can be defined via different methods, including moment-based and distance-based methods. Moment-based methods \cite{scarf1957min,delage2010} consider all the distributions whose moments satisfy certain constraints. Distance-based methods consider distributions that are within a distance $\varepsilon$ from a nominal distribution $\mathbb{P}_0$, thus capturing potential perturbations of the nominal distribution. Very often $\widehat{\mathbb{P}}_N$ is used as the nominal distribution. Commonly used distances include $\phi$-divergences \cite{bental2013,bayraksan2015}  and the Wasserstein distance \cite{datadrivenDRO,rui2023}. By minimizing the worst-case risk, one is effectively minimizing the risk with respect to all $\mathbb{Q} \in \mathcal{A}$ \cite{kuhn2019wasserstein}. Thus, if $\mathbb{P} \in \mathcal{A}$, then DRO would effectively reduce the expected risk while only having access to $\widehat{\mathbb{P}}_N$. Indeed it has been shown that when using a Wasserstein ambiguity set one can obtain confidence guarantees that $\mathbb{P} \in \mathcal{A}$ \cite{kuhn2019wasserstein,datadrivenDRO,shafieezadehabadeh2015distributionally} under certain assumptions.

Numerous works attempt to leverage DRO to address the poor out-of-sample performance of some common classification models. Many of the existing efforts consider only binary classifiers \cite{shafieezadehabadeh2015distributionally,2019regularization,kuhn2019wasserstein,FACCINI2022,taskesen2020distributionally}. This is because binary classifiers can be extended to multiclass applications using either a one-vs-one (OVO) or a one-vs-all (OVA) framework. However, both of these frameworks have their shortcomings. OVO frameworks are not scalable since the number of classifiers needed grows exponentially with the number of classes $C$. Meanwhile, OVA frameworks may not be able to fully learn correlations between the different classes \cite{cramerSVM}. Additionally, class imbalance issues are greatly amplified for OVA frameworks. Other works attempt to utilize concepts from DRO to robustify multiclass classifiers \cite{chen2023,dioxin_2023,sagawa2020,wu2023}. However, those works either consider uncertainty in the features or the labels, but not both simultaneously. To the best of our knowledge there currently exist no works that present a distributionally robust (DR) multiclass classifier in the setting where both the features and the labels are uncertain.

In this paper, we derive a tractable convex reformulation for a Wasserstein DR multiclass support vector machine (WDR-MSVM) for the setting where both the data features and labels are uncertain.
Similar to \cite{shafieezadehabadeh2015distributionally,datadrivenDRO,2019regularization,kuhn2019wasserstein} we define a separable transportation cost, and use it to construct a Wasserstein ball centered around the empirical distribution $\widehat{\mathbb{P}}_N$ of the training data $\{\widehat{\boldsymbol{\xi}}_n\}_{n=1}^{N}$. We consider the Crammer-Singer (CS) multiclass loss function introduced by \cite{cramerSVM}. First, we demonstrate various properties of the loss function, including its convexity and Lipschitz continuity, as well as the fact that it constitutes an upper bound for the empirical error of a multiclass classifier.
Then, we use strong duality results from \cite{datadrivenDRO,2019regularization} to derive a tractable convex reformulation of the worst-case risk minimization problem. Our contributions are as follows:
\begin{enumerate} [\hspace{0pt}i.]
    \item We derive a tractable convex reformulation for a WDR-MSVM equipped with a type-1 Wasserstein ambiguity set for the setting where both the features and the labels are uncertain. To achieve this we do the following:
    \begin{enumerate}[\hspace{0pt}1.]
        \item We prove that the CS loss is bounded from above by a Lipschitz continuous function for all $\mathbf{x} \in \mathcal{X}$ and $\mathbf{y} \in \mathcal{Y}$, and that each of its constituents is Lipschitz continuous for all $\mathbf{x} \in \mathcal{X}$.
        \item We leverage the strong duality results from \cite{2019regularization,datadrivenDRO} to write a convex, tractable reformulation for the linear WDR-MSVM problem.
    \end{enumerate}
    \item We also derive a tractable convex upper bound for the kernel version of the WDR-MSVM to address the setting where the classes are not linearly separable. We do this by posing the nonlinear separations between classes as hypotheses that belong to a reproducing kernel Hilbert space (RKHS). We then utilize the results in \cite{2019regularization} to show that we only need to know the kernel function to solve for the nonlinear hypotheses in a lifted learning problem without sacrificing optimality.
    \item We propose a projected subgradient method algorithm to solve a specific case of our linear WDR-MSVM training problem in a scalable fashion and we analyze its theoretical time complexity.
    \item We examine the performance of our proposed model through extensive experimentation using both simulated and real-world data. First, we explore the performance differences between the WDR-MSVM and a Wasserstein DR-OVA-SVM through simulation experiments. To that end, we demonstrate empirically that our model largely outperforms the DR-OVA-SVM when the training data is imbalanced. We then compare the performance of our model to other models from the literature using several popular real-world datasets. 
\end{enumerate}
All proofs of theoretical results are included in Appendix A.

\section{Related Work} \label{sec:related_work}
Many works investigate the use of DRO to improve the poor out-of-sample performance of common classification models. Such works propose two distinct types of models: binary ones and multiclass ones. Most of the works considering binary classifiers in the literature anticipate uncertainty both in the features and the labels. The first example of such works is \cite{2019regularization}, where the authors derive a generic tractable convex reformulation for Wasserstein  DR  binary linear classifiers. The reformulation takes different forms depending on the support of the data. Utilizing different loss functions in this formulation gives rise to different  DR  classifiers. The authors explicitly provide the  DR  reformulations for a binary SVM equipped with either the classical or the smooth hinge loss, and logistic regression. They reach their formulation by exploiting a strong dual to the worst-case risk. They then leverage the assumed regularity of the loss function to arrive at their final results. The derivation of the same Wasserstein  DR  logistic regression model had been previously presented in \cite{shafieezadehabadeh2015distributionally}, where the authors also provide out-of-sample performance guarantees for their proposed model. More specifically, they show that the Wasserstein ball acts as a confidence interval for $\mathbb{P}$ under the assumption that $\mathbb{P}$ is light-tailed. A similar Wasserstein DR binary logistic regression model is developed in \cite{taskesen2020distributionally}. However, this one incorporates an unfairness penalty that ensures that the classifier does not discriminate against sensitive, and possibly imbalanced attributes such as race, gender, or ethnicity. An alternative approach is proposed by \cite{FACCINI2022}, where the authors utilize a moment-based ambiguity set and consider the binary SVM model proposed by \cite{potra2009}. They derive a tractable reformulation for the DR SVM. They do that by projecting the ambiguity set and using duality. 
They illustrate the effectiveness in improving out-of-sample performance when compared to its deterministic counterpart through numerical experiments. However, they suggest that they are unable to derive out-of-sample performance guarantees due to their use of a moment-based ambiguity set.

On the contrary, the works that study DR multiclass classifiers often do not consider uncertainty in the features and labels simultaneously. For example, in \cite{sagawa2020}, the authors utilize group DRO to address the tendency of deep neural networks (DNNs) to learn spurious correlations in the data. They define spurious correlations as those that may manifest in the data but are not indicative of class. They develop a training algorithm for their framework and demonstrate its effectiveness in improving minority group accuracy at the cost of sacrificing some average accuracy. Nonetheless, their framework requires that any spurious correlations be known by the modeler and that the data be grouped manually based on those correlations. This drawback is addressed in \cite{wu2023}, where the authors utilize a DNN to detect any spurious correlations and group the data accordingly. However, both works still only address one failure mode of DNNs. For example, they do not consider noisy features, incorrect labels, or other overfitting modes. Moreover, both works implicitly assume that the labels are deterministic. Similarly, a Wasserstein DR multiclass logistic regression model is introduced in \cite{chen2023}. The model is very similar in derivation to the other Wasserstein  DR  classifiers presented in \cite{shafieezadehabadeh2015distributionally,2019regularization}, however, it also implicitly assumes that there is no uncertainty in the labels as the parameter controlling label flipping cost is set to $\infty$. This indicates that the distance between two samples with different labels is infinite, and therefore they cannot coexist in a Wasserstein ball with a finite radius. Finally, a label-DR multiclass classifier is introduced in \cite{dioxin_2023}. In this work, the authors focus on uncertainty in the labels and do not consider uncertainty in the features. They utilize a KL-divergence ambiguity set to hedge against the uncertainty in the labels, resulting in a family of loss functions referred to as label distributionally robust (LDR) losses.

We emphasize that our model differs fundamentally from all the works present in the literature since it is a DR multiclass classification model that hedges against uncertainty both in the features and the labels. The models introduced in \cite{shafieezadehabadeh2015distributionally,2019regularization} are perhaps the most similar to our proposed WDR-MSVM. This is due to our use of the same Wasserstein ambiguity set as the one present in those works, which leads to similarities in the derivation. However, our consideration of the multiclass CS loss introduces key differences from those works both theoretically and in performance.

\section{Problem Setup and Preliminaries} \label{sec:prob_setup}
\textbf{Multiclass Classification:} In this work, we tackle the problem of classifying data of the form $\boldsymbol{\xi} = (\mathbf{x},\mathbf{y})$, where $\mathbf{x} \in \mathcal{X} \subseteq \mathbb{R}^P$ is the input feature vector with $P$ features, and $\mathbf{y} \in \mathcal{Y} \subset \mathbb{R}^C$ is the label vector with
\begin{equation*} 
    \mathcal{Y} = \left \{ \mathbf{y} : \mathbf{y} \in \mathbb{R}^C, \ \sum_{c=1}^{C} \mathbf{y}_c = 1, \ \mathbf{y}_c \in \{ 0, \ 1\} \  \forall c \in [C] \right \},
\end{equation*}
where $C > 2$ is the number of classes. We denote the support set of the data by $\Xi = \mathcal{X} \times \mathcal{Y}$. Suppose we have access to a training dataset $\mathcal{T}=\{\widehat{\boldsymbol{\xi}}_{(n)}\}_{n=1}^{N} = \{(\widehat{\mathbf{x}}_{(n)},\widehat{\mathbf{y}}_{(n)})\}_{n=1}^N$ comprised of $N$ training samples. As discussed in \cite{cramerSVM}, a multiclass classifier $H_{\mathbf{M}} \colon \mathcal{X} \rightarrow \mathcal{Y}$ parameterized by $\mathbf{M} \in \mathcal{M} =  \mathbb{R}^{C \times P}$ seeks to map an instance of the features $\widehat{\mathbf{x}}_{(n)}$ to a predicted label vector $\mathbf{y}^{\ast}_{(n)}$. The output of such a classifier would be
\begin{align*} 
     H_{\mathbf{M}}(\widehat{\mathbf{x}}_n)_c &= \mathbf{y}^{\ast}_{(n)_c}\\
     &= \left\{ \begin{array}{ll} 1 & \text{if } c = \argmax_{j \in [C]}\mathbf{M}_{j \mathbf{\cdot}} \mathbf{x} \\  0 & \text{Otherwise}
  \end{array} \right. \forall c \in [C].
\end{align*} 
Therefore, a sample is correctly classified if $\mathbf{y}^{\ast}_{(n)} = \widehat{\mathbf{y}}_{(n)}$, and is misclassified otherwise. Thus, one can compute the empirical error $\epsilon(H_{\mathbf{M}})$ of classifier $H_{\mathbf{M}}$ via
\begin{equation*} \label{eq:empirical_error}
    \epsilon(H_{\mathbf{M}}) = \frac{1}{N} \sum_{n=1}^{N} \mathbf{1}_{ \left \{ H_{\mathbf{M}} \left ( \widehat{\mathbf{x}}_{(n)} \right ) \neq \widehat{\mathbf{y}}_{(n)} \right \} }.
\end{equation*}
While evaluating the empirical error of an existing classifier is straightforward, solving the minimization problem of the discrete empirical error $\epsilon(H_{\mathbf{M}})$ to obtain optimal model parameters $\mathbf{M}^{\ast}$ is computationally expensive \cite{cramerSVM,crammer2002}. To address this, \cite{cramerSVM} derive a continuous, piecewise linear approximation of $\epsilon(H_{\mathbf{M}})$. This approximation is shown next and is often referred to as the CS loss.

\begin{definition}[\cite{cramerSVM}] \label{def:cs_loss}
    The CS loss $\ell_{\text{CS}}(\mathbf{M};\boldsymbol{\xi})$ parameterized by $\mathbf{M} \in \mathcal{M}$ is defined as
    \begin{equation*} \label{eq:cs_loss}
    \begin{aligned}
        \ell_{\text{CS}} (\mathbf{M};\boldsymbol{\xi}) &\coloneqq   \max_{c \in [C]} \left \{ \left ( \mathbf{v}_{(c)} \right )^\mathsf{T} \left ( \mathbf{M} \mathbf{x} - \mathbf{y} \right ) + 1 \right \} - \mathbf{y}^{\mathsf{T}} \mathbf{M} \mathbf{x},
    \end{aligned}
    \end{equation*}
    where $\mathbf{v}_{(c)} \in \mathcal{Y}$ is such that $\mathbf{v}_{(c)_c}=1$.
\end{definition}
\begin{lemma} \label{lem:prop_loss}
    The CS loss $\ell_{\text{CS}}(\mathbf{M};\boldsymbol{\xi}) = \ell_{\text{CS}}(\mathbf{M};(\mathbf{x},\mathbf{y})))$ defined in Def. \ref{def:cs_loss} possesses the following properties:
    \begin{enumerate} [\hspace{0pt}i.]
        \item $ \frac{1}{N} \sum_{n=1}^{N} \ell_{\text{CS}}(\mathbf{M};(\widehat{\mathbf{x}}_{(n)},\widehat{\mathbf{y}}_{(n)})) \geq \epsilon(H_{\mathbf{M}}) \forall \mathbf{M} \in \mathcal{M}$
        \item\label{part:bounded_above} $\ell_{\text{CS}}(\mathbf{M};(\mathbf{x},\mathbf{y}))$ is bounded from above by a function $f((\mathbf{x},\mathbf{y}))$ that is Lipschitz continuous in $\mathbf{x}$ and $\mathbf{y}$ for all $\mathbf{M} \in \mathcal{M}$
        \item $\ell_{\text{CS}}(\mathbf{M};(\mathbf{x},\mathbf{y}))$ is convex in $\mathbf{x}$ for all $\mathbf{M} \in \mathcal{M}$, $\mathbf{y} \in \mathcal{Y}$
        \item $\ell_{\text{CS}}(\mathbf{M};(\mathbf{x},\mathbf{y}))$ is convex in $\mathbf{M}$ for all $\mathbf{x} \in \mathcal{X}$, $\mathbf{y} \in \mathcal{Y}$
        \item\label{part:lip_mod} Each individual constituent $\ell_{\text{CS},c}(\mathbf{M};(\mathbf{x},\widecheck{\mathbf{y}})) = \left ( \mathbf{v}_{(c)} \right )^{\mathsf{T}}(\mathbf{M} \mathbf{x} - \widecheck{\mathbf{y}}) + 1 - \widecheck{\mathbf{y}}^{\mathsf{T}}\mathbf{M} \mathbf{x}$ is Lipschitz continuous in $\mathbf{x}$ for all $\mathbf{M} \in \mathcal{M}$ given a fixed $\widecheck{\mathbf{y}} \in \mathcal{Y}$ with Lipschitz modulus $\Lip \left (\ell_{\text{CS},c}(\mathbf{M};(\mathbf{x},\widecheck{\mathbf{y}})) \right ) = \left | \left | (\mathbf{v}_{(c)} - \widecheck{\mathbf{y}})^{\mathsf{T}} \mathbf{M}\right | \right |_{\ast}$, where $||\cdot||_{\ast}$ is the dual of the norm used to measure distances between instances of $\mathbf{x}$
    \end{enumerate}
\end{lemma}
Therefore, the CS loss obeys the regularity assumptions needed to derive tractable convex formulations for Wasserstein DRO programs \cite{datadrivenDRO,kuhn2019wasserstein,2019regularization}. 

\subsection{Wasserstein Distributionally Robust Optimization:} Distributionally robust optimization aims to hedge against the uncertainty in cases where it is governed by an unknown distribution $\mathbb{P}$. It is an approach that can be viewed as a middle-ground between SP \cite{shapiro_stochprog} and robust optimization (RO) \cite{bental_ro}. Indeed DRO cannot characterize the uncertainty as well as SP since $\mathbb{P}$ is unknown. However, it is not as overly conservative as RO can often be \cite{bertsimas2004}. As mentioned previously, DRO aims to minimize the worst-case risk achieved by any distribution that belongs to an ambiguity set $\mathcal{A}$. This problem is mathematically formulated as
\begin{equation} \label{eq:dro_problem}
    \inf_{\ell \in \mathcal{L}} \sup_{\mathbb{Q} \in \mathcal{A}} \mathbb{E}^{\mathbb{Q}}[\ell],
\end{equation}
where $\mathcal{L}$ is the set of all loss functions being considered. Tractable reformulations for this problem in various generic cases of interest are derived in many works such as \cite{datadrivenDRO,kuhn2019wasserstein,bayraksan2015,shapiro2017,rui2023}. In our work we utilize DRO equipped with a Wasserstein ambiguity set due to its many attractive properties discussed in \cite{kuhn2019wasserstein,rui2023}. Such properties include the ambiguity set's ability to contain both continuous and discrete distributions regardless of the structure of the nominal distribution $\mathbb{P}_0$, and the ability to derive out-of-sample performance guarantees.

The Wasserstein ambiguity set $\mathcal{A}_{\varepsilon,q}(\Xi)$ is defined as a ball of radius $\varepsilon$ in the sense of the type-$q$ Wasserstein distance centered at a distribution $\widehat{\mathbb{P}}_N$ estimated from the data in $\mathcal{T}$. 
\begin{definition}[\cite{kant1960}] \label{def:wass_distance}
    The type-$q$ Wasserstein distance between two distributions $\mathbb{Q}$ and $\mathbb{Q}^{\prime}$ represents the the minimum cost of transforming $\mathbb{Q}$ to $\mathbb{Q}^{\prime}$, and is defined as
    \begin{equation*}\label{eq:wass_distance}
    W_{d,q}(\mathbb{Q},\mathbb{Q}') \coloneqq \left ( \inf_{\pi \in \Pi(\mathbb{Q},\mathbb{Q}')} \int_{\Xi \times \Xi} d \left( \boldsymbol{\xi}, \boldsymbol{\xi}^{\prime} \right )^q \pi(\text{d} \boldsymbol{\xi}, \text{d} \boldsymbol{\xi}^{\prime}) \right )^{\frac{1}{q}},
    \end{equation*}
    where $d \left( \boldsymbol{\xi}, \boldsymbol{\xi}^{\prime} \right )$ denotes the transportation cost per unit mass from $\boldsymbol{\xi}$ to $\boldsymbol{\xi}^{\prime}$, and $\Pi(\mathbb{Q},\mathbb{Q}^{\prime})$ is the set of all joint distributions of $\boldsymbol{\xi}$ and $\boldsymbol{\xi}^{\prime}$ with marginals $\mathbb{Q}$ and $\mathbb{Q}^{\prime}$, respectively.
\end{definition}

Thus, the Wasserstein ambiguity set $\mathcal{A}_{\varepsilon,q}(\Xi)$ is written as
\begin{equation*}
    \mathcal{A}_{\varepsilon,q}(\Xi) \coloneqq \left \{ \mathbb{Q} \in \mathcal{P} \left ( \Xi \right ) \colon W_{d,q} \left ( \mathbb{Q},\widehat{\mathbb{P}}_N\right ) \leq \varepsilon \right \},
\end{equation*}
where $\mathcal{P}(\Xi)$ is the set of all distributions supported on $\Xi$. 
\section{Wasserstein Distributionally Robust Multiclass SVM} \label{sec:DR-SVM}

\textbf{Linear WDR-MSVM:} In constructing our WDR-MSVM, we set $\widehat{\mathbb{P}}_N$ as the empirical distribution of the training data from $\mathcal{T}$. Additionally, we restrict our focus to an ambiguity set $\mathcal{A}_{\varepsilon,1}(\Xi)$ defined via the type-$1$ Wasserstein distance. Moreover, we utilize the separable transportation cost function from \cite{shafieezadehabadeh2015distributionally,2019regularization}, which is
\begin{equation} \label{eq:trans_cost}
    d \left( \boldsymbol{\xi}, \boldsymbol{\xi}^{\prime} \right ) \coloneqq ||\mathbf{x} - \mathbf{x}^{\prime}|| + \kappa \mathbf{1}_{ \{\mathbf{y} \neq \mathbf{y}^{\prime} \}},
\end{equation}
where $||\cdot||$ is any norm on $\mathbb{R}^P$, and $\kappa$ is a user parameter that can be viewed as the distance between two samples with identical features and different labels. Alternatively, if one were to view the radius $\varepsilon$ as the maximum total budget to be spent on perturbing $\widehat{\mathbb{P}}_N$, then $\kappa$ can be viewed as the cost of changing a sample label. This separable cost function offers two advantages. Firstly, the parameter $\kappa$ makes the model more flexible as it allows to characterize different levels of uncertainty in the labels. Secondly, the separability of the cost function combined with the finite number of classes enables the separation of the dual optimization problem with respect to $\mathcal{X}$ and $\mathcal{Y}$. This will be key to dealing with the non-convexity of $\mathcal{Y}$ and deriving tractable reformulations for the proposed WDR-MSVM.
\begin{assumption}\label{assump:feature_support}
    The support set $\mathcal{X}$ of the features $\mathbf{x}$ is the entire space (i.e. $\mathcal{X} = \mathbb{R}^P$).
\end{assumption}

We note that this is not a restrictive assumption. Indeed, our proposed model can still obtain feasible solutions, albeit sub-optimal, in cases where the features are continuous and $\mathcal{X} \subset \mathbb{R}^P$. To see this, note that $\mathcal{A}_{\varepsilon,1}(\mathcal{X} \times \mathcal{Y}) \subseteq \mathcal{A}_{\varepsilon,1}(\mathbb{R}^P \times \mathcal{Y})$ whenever $\mathcal{X} \subset \mathbb{R}^P$. Therefore, $\sup_{\mathbb{Q} \in \mathcal{A}_{\varepsilon,1}(\mathcal{X} \times \mathcal{Y})} \mathbb{E}^{\mathbb{Q}}[\ell] \leq \sup_{\mathbb{Q} \in \mathcal{A}_{\varepsilon,1}(\mathbb{R}^P \times \mathcal{Y})} \mathbb{E}^{\mathbb{Q}}[\ell]$. Given the previously stated conditions, we derive the following tractable convex reformulation for the WDR-MSVM.
\begin{theorem}\label{thm:linear_DRMSVM}
    Under the condition of Asm. \ref{assump:feature_support}, if the type-$1$ Wasserstein distance equipped with the transportation cost defined in \eqref{eq:trans_cost} is used, and the ambiguity set is centered at the empirical distribution $\widehat{\mathbb{P}}_N$ of the training data in $\mathcal{T}$, then the DRO problem in \eqref{eq:dro_problem} for the CS loss $\ell_{\text{CS}}(\mathbf{M};\boldsymbol{\xi})$ defined in \eqref{eq:cs_loss} admits the following tractable convex reformulation
    \begin{equation*} \label{eq:wdro_msvm}
    \begin{aligned}
        &\inf_{\mathbf{M}} \sup_{\mathbb{Q} \in \mathcal{A}_{\varepsilon,1}(\Xi)} \mathbb{E}^{\mathbb{Q}} \left [ \ell_{\text{CS}}(\mathbf{M};\boldsymbol{\xi}) \right ] =\\
        &\left \{\begin{aligned}
            &\min_{\mathbf{M},\lambda,s_n} && \lambda \varepsilon + \frac{1}{N} \sum_{n=1}^{N} s_n&& \\
            & \subjectto && \ell_{\text{CS}} (\mathbf{M};(\widehat{\mathbf{x}}_{(n)},\widehat{\mathbf{y}}_{(n)})) \leq s_n && \forall n \in [N] \\
            &&& \ell_{\text{CS}} (\mathbf{M};(\widehat{\mathbf{x}}_{(n)},\widecheck{\mathbf{y}}_{(c)})) - \lambda \kappa \leq s_n && \forall n \in [N],\\
            &&&&& \forall \widecheck{\mathbf{y}}_{(c)} \in \mathcal{Y},\\
            &&&&&\widecheck{\mathbf{y}}_{(c)} \neq \widehat{\mathbf{y}}_{(n)}\\
            &&& \lambda \geq \left | \left | (\mathbf{v}_{(i)} - \mathbf{v}_{(j)})^{\mathsf{T}}\mathbf{M} \right | \right |_{\ast} && \forall i,j \in [C]
        \end{aligned} \right.
    \end{aligned}
    \end{equation*}
    where $||\cdot||_{\ast}$ is the dual to the norm used in the transportation cost function in \eqref{eq:trans_cost}.
\end{theorem}

\begin{remark}
    Consider the special case of the formulation presented in \eqref{eq:wdro_msvm} where $\kappa = \infty$. In this case, the second group of constraints disappears from the formulation, leaving $\lambda$ only in the objective function and the final group of constraints. Therefore, at optimality we get $\lambda^{\ast} = \max_{(i,j) \in [C]}\left | \left | (\mathbf{v}_{(i)} - \mathbf{v}_{(j)})^{\mathsf{T}}\mathbf{M} \right | \right |_{\ast}$. Thus, the DR model reduces to a regularized one with regularizer $R(\mathbf{M}) =  \lambda^{\ast}$ and regularization parameter $\varepsilon$. This indicates that the regularized formulation (hereinafter referred to as R-MSVM) is a special case of the type-$1$ Wasserstein DR formulation where the labels are assumed certain. Thus, we would expect the WDR-MSVM to outperform it in most real-world settings as discussed in  \cite{shafieezadehabadeh2015distributionally}.
\end{remark}

\textbf{Kernel WDR-MSVM:} Now, we extend our model to consider nonlinear class separation represented via hypotheses $\mathbf{h}$ that belong to a reproducing kernel Hilbert space (RKHS) $\mathbb{H} \subseteq \mathbb{R}^{\mathcal{X}}$. The RKHS is equipped with a self-dual norm $||\cdot||_{\mathbb{H}}$ and kernel function $k(\cdot,\cdot)$, which is assumed to satisfy the following calmness assumption discussed in \cite{2019regularization}.

\begin{assumption}[\cite{2019regularization}]\label{assump:calm}
    The kernel function $k$ used to define the RKHS $\mathbb{H}$ is calm from above. That is, there exists a concave smooth growth function $f:\mathbb{R}_+ \rightarrow \mathbb{R}_+$ with $f(0) = 0$ and $f^{\prime}(z) \geq 1 \ \forall z \in \mathbb{R}_+$ such that 
    \begin{equation*}
    \sqrt{k(\mathbf{x}_{(1)},\mathbf{x}_{(1)}) - 2k(\mathbf{x}_{(1)},\mathbf{x}_{(2)}) + k(\mathbf{x}_{(2)},\mathbf{x}_{(2)})} \leq f(||\mathbf{x}_{(1)} - \mathbf{x}_{(2)} ||_2) \qquad \forall \mathbf{x}_{(1)},\mathbf{x}_{(2)} \in \mathcal{X}.
    \end{equation*}
\end{assumption}
Note that it is shown in \cite{2019regularization} that many commonly used kernels satisfy Asm. \ref{assump:calm}. Leveraging the kernel function $k(\cdot,\cdot)$, we derive the following tractable convex upper bound for the kernel WDR-MSVM.

\begin{theorem} \label{thm:kernel_DRMSVM}
    Suppose that $||\cdot||_{\mathbb{H}} = ||\cdot||_2$, and that all definitions and assumptions from this subsection hold. Then, we can write the following tractable convex upper bound for the kernel version of the WDR-MSVM as follows
    \begin{equation*}
    \begin{aligned}
        &\inf_{\{ \mathbf{h}_{(c)}\}_{c=1}^C \subset \mathbb{H}} \sup_{\mathbb{Q} \in \mathcal{A}_{\varepsilon,1}^{\mathbb{H}}(\Xi)} \mathbb{E}^{\mathbb{Q}} [\ell_{\text{CS},\mathbb{H}}(\mathbf{g}_{(\mathbb{H})}(\mathbf{x}_{(\mathbb{H})});\mathbf{y})] = \\
        &\left \{ \begin{aligned}
            & \min_{\mathbf{A},\lambda,s_n} && \lambda \varepsilon + \frac{1}{N} \sum_{n=1}^{N} s_n&& \\
            & \subjectto && \max_{c \in [C]} \left \{ \mathbf{v}_{(c)}^{\mathsf{T}} \left ( \sum_{j=1}^{N} \mathbf{A}_{\cdot j}\mathbf{K}_{nj} - \widehat{\mathbf{y}}_{(n)} \right ) + 1 \right \} - \widehat{\mathbf{y}}_{(n)}^\mathsf{T}\sum_{j=1}^{N} \mathbf{A}_{\cdot j}\mathbf{K}_{nj} \leq s_n &&\forall n \in [N] \\
            &&& \max_{c \in [C]} \Bigg \{ \mathbf{v}_{(c)}^{\mathsf{T}} \bigg ( \sum_{j=1}^{N} \mathbf{A}_{\cdot j}\mathbf{K}_{nj} - \widecheck{\mathbf{y}}_{(i)} \bigg ) + 1 \bigg \} - \widecheck{\mathbf{y}}_{(i)}^\mathsf{T}\sum_{j=1}^{N} \mathbf{A}_{\cdot j}\mathbf{K}_{nj} - \lambda \kappa \leq s_n &&\forall n \in [N],\forall \widecheck{\mathbf{y}}_{(i)} \in \mathcal{Y}, \ \widecheck{\mathbf{y}}_{(i)} \neq \widehat{\mathbf{y}}_{(n)}\\
            &&& \lambda \geq \left | \left | \mathbf{K}^{\frac{1}{2}} \left ( \mathbf{A}_{r \cdot}\right )^{\mathsf{T}} \right | \right |_2 + \left | \left | \mathbf{K}^{\frac{1}{2}} \left ( \mathbf{A}_{s \cdot}\right )^{\mathsf{T}} \right | \right |_2 &&\forall r,s \in [C]
        \end{aligned} \right.
    \end{aligned}
    \end{equation*}
    where $\mathbf{K}_{ij} = k(\widehat{\mathbf{x}}_{(i)},\widehat{\mathbf{x}}_{(j)})$, and $\mathbf{A} \in \mathbb{R}^{C \times N}$ is such that the index of the largest element in $\sum_{n=1}^N \mathbf{A}_{\cdot n} k(\mathbf{x}_s,\widehat{\mathbf{x}}_n)$ is the predicted class of test sample $\mathbf{x}_s$.

\end{theorem}

\textbf{Solution Algorithm}
 Observe that the linear WDR-MSVM problem in Thm. \ref{thm:linear_DRMSVM} can be written as a linear program (LP) if the $\ell_1$ or the $\ell_{\infty}$-norm are used in the transportation cost function \eqref{eq:trans_cost}, and as a quadratically constrained quadratic program (QCQP) when the $\ell_2$-norm is used in \eqref{eq:trans_cost}. Similarly, the kernel WDR-MSVM problem in Thm. \ref{thm:kernel_DRMSVM} can be written as a QCQP. Therefore, both problems are solvable via the barrier method.
 While this suggests that both problems can be solved via off-the-shelf solvers such as Gurobi, the theoretical worst case time complexity of the problems is undesirably high as shown in Appendix B. In the following, we propose a scalable algorithm to solve a specific case of the linear WDR-MSVM problem in Thm. \ref{thm:linear_DRMSVM}.

\begin{theorem}\label{thm:alg}
 Suppose the $\ell_{\infty}$-norm is used in the linear WDR-MSVM problem in Thm. \ref{thm:linear_DRMSVM}. Then, the problem is solvable via the following projected subgradient method algorithm.

\begin{algorithm}[H]
    \caption{Projected Subgradient Method Algorithm}
    \label{alg:subgrad}
    \small
    \textbf{Input}: $\mathbf{M}^{(0)}$, $\lambda^{(0)}$\\
    \textbf{Parameter}: Number of iterations $T$, stepsize $\sigma(t)$ at $t^{th}$ iteration\\
    \textbf{Output}: $\mathbf{M}^{\ast}$, $\lambda^{\ast}$
    \begin{algorithmic}[1] 
        \FOR{$t = 1, \dots, T$}
        \STATE $\lambda^{\prime} \leftarrow \lambda^{(t)} - \sigma(t) \left ( \varepsilon + \kappa \sum_{n=1}^{N} \mathbf{1}_{\{\tau(\lambda,\mathbf{M})\}}(n) \right )$
        \STATE $\mathbf{v}_{(c)}^{\ast}(n),\mathbf{y}_{(c)}^{\ast}(n) \leftarrow \argmax_{\mathbf{v}_{(c)}\in \mathcal{Y},\mathbf{y}_{(c)} \in \mathcal{Y}} L(n)$
        \STATE $\mathbf{M}^{\prime} \leftarrow \mathbf{M}^{(t)} - \sigma(t) \sum_{n=1}^{N} \left (\mathbf{v}_{(c)}^{\ast}(n) - \mathbf{y}_{(c)}^{\ast}(n) \right )^{\mathsf{T}} \widehat{\mathbf{x}}_{(n)}$
        \STATE $\lambda^{(t+1)},\mathbf{M}^{(t+1)} \leftarrow \argmin_{\lambda,\mathbf{M}} \Pi(\lambda^{\prime},\mathbf{M}^{\prime})$
        \ENDFOR
    \end{algorithmic}
\end{algorithm}
where $\mathbf{1}_{\{\tau(\lambda,\mathbf{M})\}}(n)$ is equivalent to $1$ if $\ell_{\text{CS}} (\mathbf{M};(\widehat{\mathbf{x}}_{(n)},\widecheck{\mathbf{y}}_{(c)})) - \lambda \kappa > \ell_{\text{CS}} (\mathbf{M};(\widehat{\mathbf{x}}_{(n)},\widehat{\mathbf{y}}_{(n)}))$ for any $\widecheck{\mathbf{y}}_{(c)} \in \mathcal{Y}, \widecheck{\mathbf{y}}_{(c)} \neq \widehat{\mathbf{y}}_{(n)}$, and is equivalent to $0$ otherwise, $L(n) =  \left (\mathbf{v}_{(c)} \right )^\mathsf{T} \left ( \mathbf{M} \widehat{\mathbf{x}}_n - \mathbf{y}_{(c)} \right ) - \mathbf{y}_{(c)}^{\mathsf{T}} \mathbf{M} \widehat{\mathbf{x}}_n - \lambda \kappa \mathbf{1}_{\{\mathbf{y}_{(c)} \neq \widehat{\mathbf{y}}_n\}}$, and $\Pi(\lambda^{\prime},\mathbf{M}^{\prime})$ is the following projection problem:
\begin{equation}
\nonumber \Pi(\lambda^{\prime},\mathbf{M}^{\prime}) = \left \{    \begin{aligned}
    &\min_{\substack{\lambda,\mathbf{M},\\ m_d}} &&(\lambda - \lambda^{\prime})^2 + \sum_{c=1}^{C} \sum_{p = 1}^{P} (\mathbf{M}_{cp} - \mathbf{M}_{cp}^{\prime})^2\\
    &\subjectto && m_d - \frac{\lambda}{2} \leq \mathbf{M}_{id} \leq m_d + \frac{\lambda}{2} \forall i \in [C], \forall d \in [P]
\end{aligned} \right.
\end{equation}
\end{theorem}

\begin{proposition}\label{prop:conv}
    Suppose that the simplified projection problem $\Pi(\lambda^{\prime},\mathbf{M}^{\prime})$ in Thm. \ref{thm:alg} is solved via the barrier method equipped with the log barrier function and Newton updates. Moreover, suppose that $\epsilon_1$ and $\epsilon_2$ are the optimality tolerances of the linear WDR-MSVM problem in Thm. \ref{thm:linear_DRMSVM} and the projection problem $\Pi(\lambda^{\prime},\mathbf{M}^{\prime})$, respectively. Then Alg. \ref{alg:subgrad} has a theoretical worst-case time complexity of $\mathcal{O} \left (\epsilon_1^{-2} \left [ NPC^2 + NC^3 + P^{3.5}C^{3.5} \log(\beta \epsilon_2^{-1}) \right ] \right )$.
\end{proposition}
\begin{remark}
    Alg. \ref{alg:subgrad} proposed offers the following advantages over the barrier method in off-the-shelf solvers:
    \begin{enumerate}
        \item It enjoys a more scalable worst-case time complexity.
        \item It is susceptible to the use of stochastic subgradient approaches to improve scalability for cases with large $N$.
    \end{enumerate}
\end{remark}

\section{Numerical Experiments} \label{sec:exp}
All the error bars on the plots represent one standard deviation. Additional experimental results (including scalability experiments) and software, hardware, and dataset information are included in Appendices C and D, respectively.
\subsection{Experiment 1: Simulation Sensitivity Analysis}
The goal of the simulation experiment is to study the differences in performance between our proposed linear WDR-MSVM and the DR-OVA-SVM from \cite{2019regularization} in a controlled setting. We generate data via the \texttt{make\_classification} module from the \texttt{scikit-learn} Python package \cite{scikit-learn}. The data is grouped into $C \in \{4,8\}$ classes. Each class is located at a vertex of a $P$-dimensional hypercube, where $P \in \{3,5,15,30\}$ is the number of features. The data at each vertex $c \in [C]$ is sampled from a Gaussian distribution with variance $\sigma^2_c=1$ and mean $\mu_c$ coinciding with the vertex location. The separation between the vertices is fixed at $3$. We utilize $N=200$ training samples and $N_{\text{test}} = 2000$ testing samples for all runs. We test two conditions for each combination of $C$ and $P$: balanced and imbalanced training set. The distribution of the data across classes in the imbalanced setting is $(45\%,25\%,25\%,5\%)$ and $(20\%,20\%,12.5\%,12.5\%,12.5\%,12.5\%,5\%,5\%)$ in runs where $C=4$ and $C=8$, respectively. The test set is balanced for all runs. Finally, each experimental combination is repeated $50$ times with randomly generated datasets for each repetition. During each experiment, we sweep over values of $\varepsilon \in [1\times10^{-6},1\times10^1]$ and $\kappa \in [0,1]$ for both models. We utilize the $\ell_{\infty}$-norm in the transportation cost \eqref{eq:trans_cost} for all models. We evaluate the mean correct classification rate (mCCR) over the test set at each parameter combination to examine the out-of-sample performance of the models.

\begin{figure*}[t]
\centering
\begin{subfigure}{1\textwidth}
    \includegraphics[width=\textwidth]{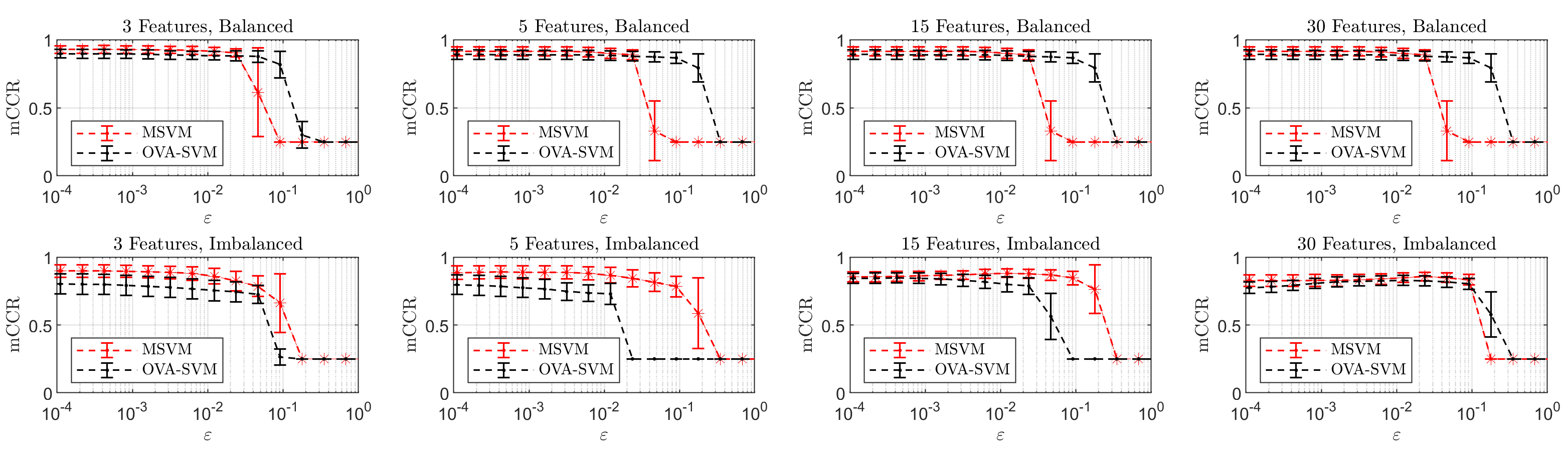}
    \caption{Plots of mCCR vs. $\varepsilon$ for the simulation experiment with $4$ classes.}
    \label{fig:4classes_sim}
\end{subfigure}
\hfill
\begin{subfigure}{1\textwidth}
    \includegraphics[width=\textwidth]{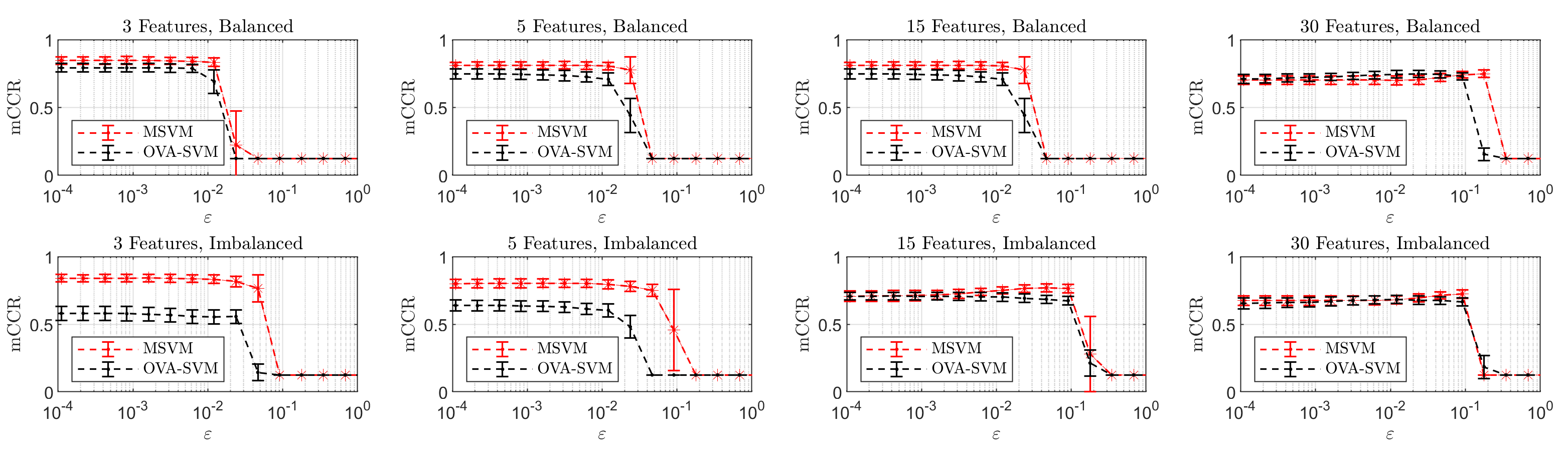}
    \caption{Plots of mCCR vs. $\varepsilon$ for the simulation experiment with $8$ classes.}
    \label{fig:8classes_sim}
\end{subfigure}
\caption{Results of the simulation experiments.}
\end{figure*}

The plots of mCCR vs. $\varepsilon$ at the value of $\kappa$ that attains peak mCCR are displayed in Figs. \ref{fig:4classes_sim} and \ref{fig:8classes_sim} for $C=4$ and $C=8$, respectively. We observe that when the data is balanced, the WDR-MSVM model offers a modest improvement of $0.02\%$ to $5.40\%$ over the OVA model in terms of peak mCCR. However, the OVA model maintains its peak mCCR for a wider range of $\varepsilon$. This makes it a more attractive model in practice due to its lower sensitivity to the hyperparameter value. However, we observe that in the setting of imbalanced data when $P$ is less than $C$, the WDR-MSVM model offers a much more substantial improvement of $2.86\%$ to $25.98\%$ in peak mCCR over its OVA counterpart. This showcases the fact that OVA frameworks often struggle in settings with imbalanced training data. Surprisingly however, the advantage of the WDR-MSVM diminishes as $P$ exceeds $C$, where both models achieve similar performance. This suggests that our proposed model would ideally be applied in settings where the training data is known to exhibit high imbalance, and the number of features does not largely exceed that of the classes. An example of such applications is fault diagnostics in industrial settings given limited sensor data. In this case, many training samples may be available for a healthy system state, but very few samples can be obtained for the different faults before the system is repaired. Similarly, our model could be used in healthcare applications for the diagnosis of a rare disease. 

\subsection{Experiment 2: Real-World Experiment}

\begin{table*}
\small
\centering
\begin{tabular}{l|lllll}
             & \multicolumn{1}{c}{Wine}                       & \multicolumn{1}{c}{Penguins}          & \multicolumn{1}{c}{Iris}              & \multicolumn{1}{c}{Seeds}             & \multicolumn{1}{c}{AI4I}              \\ \hline
DR-MSVM      & {\textbf{95.92\%±2.33\%}} & {\textbf{99.25\%±0.19\%}} & {95.6\%±2.31\%}  & {96.00\%±2.29\%} & {67.27\%±5.22\%} \\
DR-OVA       & {95.81\%±2.20\%}          & {99.16\%±0.30\%} & {94.09\%±3.10\%} & {\textbf{96.34\%±2.57\%}} & {60.38\%±6.70\%} \\
R-MSVM       & {95.25\%±2.54\%}          & {98.35\%±0.24\%} & {95.73\%±2.28\%} & {95.48\%±2.90\%} & {69.36\%±5.40\%}\\
R-MLR &95.36\%±2.54\% &98.80\%±0.79\% &96.88\%±2.51\% &95.40\%±2.82\%	 &\textbf{69.74\%±5.63\%}\\ \hline
kDR-MSVM & {73.66\%±5.01\%}          & {83.25\%±1.33\%} & {\textbf{97.07\%±2.17\%}} & {94.20\%±3.80\%} & {44.76\%±3.05\%} \\
kDR-OVA  & {73.74\%±5.10\%}          & {83.47\%±1.51\%} & {96.89\%±2.33\%} & {94.34\%±3.65\%} & {44.60\%±3.06\%} \\
kR-MSVM  & {73.66\%±5.01\%}          & {83.25\%±1.33\%} & {96.53\%±2.51\%} & {94.17\%±3.70\%} & {44.74\%±3.05\%} \\ \hline
\end{tabular}
\captionof{table}{Peak mCCR values and standard deviation achieved by all tested models for all datasets}
\label{table:real_exp}
\end{table*}

In this section we utilize real-world datasets to compare the performance of our proposed WDR-MSVM with that of the DR-OVA-SVM, the regularized version of our proposed model (R-MSVM), and a regularized multinomial logistic regression model (R-MLR). We compare the linear and RBF kernel versions of the three SVM models. We utilize 5 popular datasets from the UCI repository in our study: Wine \cite{misc_wine_109}, Seeds \cite{misc_seeds_236}, Palmer Penguins \cite{misc_penguins}, Iris \cite{misc_iris_53}, and AI4I \cite{misc_ai4i_2020_predictive_maintenance_dataset_601}. We use 70\% of the dataset as a training set and the rest as a testing set for all the datasets used except for AI4I. For AI4I we use a training set comprised of 300 samples, 92.5\% of which is healthy samples and the rest is divided equally over the fault classes. This is done to reflect the imbalance present in the dataset. The test set consists of 200 samples distributed in a balanced fashion across all classes. We train each model on the data for $\varepsilon \in \{ 1\times10^{-5},1\times10^{-4},1\times10^{-3},1\times10^{-2},1\times10^{-1}\}$, and for $\kappa \in \{ 0.25,0.5,0.75,1\}$. We utilize the $\ell_{\infty}$-norm in the transportation cost \eqref{eq:trans_cost} for the linear models. Moreover, we set the kernel parameter $\gamma$ to $\frac{1}{P}$ for the kernel models.
We repeat each experiment 50 times, and the training and test sets in each run are randomized. We then compute the mCCR for all experimental conditions, and report the max mCCR achieved by each model in Tab. \ref{table:real_exp}. 

We observe that for most datasets, the best-performing version of the WDR-MSVM outperforms that of the DR-OVA-SVM. However, for Wine, Penguins, and Iris the difference between them is very small ($0.09\%$ to $0.18\%$), and for Seeds the DR-OVA-SVM even outperforms the WDR-MSVM. This aligns with the results presented in the simulation experiments, as all four of those datasets are balanced. Thus, the two models tend to perform similarly. However, when the models are used on the highly imbalanced AI4I dataset, WDR-MSVM outperforms its OVA counterpart by a significant $6.89\%$. This emphasizes that the WDR-MSVM model excels in applications with imbalanced training data.

Moreover, we observe that the best-performing version of the WDR-MSVM outperforms that of the R-MSVM for most datasets by $0.52\%$ to $0.9\%$. Recall that the only difference between the models is that WDR-MSVM accounts for uncertain labels while the regularized version does not. This suggests that in most real-world applications accounting for label uncertainty can lead to some classifier improvement. However, we also observe that in the AI4I dataset, R-MLR and R-MSVM outperform WDR-MSVM by $2.09\%$ and $2.47\%$, respectively. We hypothesize that this is due to minimal label uncertainty in the dataset, making the WDR-MSVM overly conservative.

\section{Conclusions and Future Work} \label{sec:conc}
In this paper we proposed a Wasserstein distributionally robust multiclass SVM. We derived a tractable convex reformulation and upper bound for the linear and the kernel versions of our proposed model, respectively. We also proposed a scalable solution algorithm for a specific case of the linear WDR-MSVM. We then empirically showed through simulation experiments that our proposed model outperforms its OVA counterpart in cases where the training set is imbalanced and the number of classes exceeds that of the features.
Subsequently, we demonstrated through experiments using popular real-world datasets that linear and kernel versions of our model often outperform existing models. Future extensions would focus on developing scalable training algorithms for all versions of the WDR-MSVM, and on extending the models to problems with mixed or bounded features.


\section*{Acknowledgements}
This work was funded by the National Aeronautics and Space Administration (NASA), Space Technology Research Institute  (STRI)  Habitats  Optimized  for  Missions  of  Exploration (HOME) ‘SmartHab’ Project (Grant No. 80NSSC19K1052).

\bibliographystyle{ieeetr}  
\bibliography{references}

\newpage
\section{Appendix A: Proofs of Theoretical Results}
\subsection{Proof of Lemma \ref{lem:prop_loss}}
\textit{Proof.} Next we show the five properties stated in Lemma \ref{lem:prop_loss}. 
\begin{enumerate} [\hspace{0pt}i.]
    \item \begin{subequations}
        \begin{align}
            \epsilon(H_{\mathbf{M}}) &\label{eq:p1_1}= \frac{1}{N}\sum_{n=1}^{N} \mathnormal{1}_{\{ H_{\mathbf{M}}(\widehat{\mathbf{x}}_{(n)})\neq \widehat{\mathbf{y}}_{(n)}\}}\\
            &\label{eq:p1_2}\leq \frac{1}{N}\sum_{n=1}^{N} H_{\mathbf{M}}(\widehat{\mathbf{x}}_{(n)})^{\mathsf{T}}\mathbf{M} \widehat{\mathbf{x}}_{(n)} - \widehat{\mathbf{y}}_{(n)}^{\mathsf{T}}\mathbf{M}\widehat{\mathbf{x}}_{(n)} - H_{\mathbf{M}}(\widehat{\mathbf{x}}_{(n)})^{\mathsf{T}}\widehat{\mathbf{y}}_{(n)} + 1\\
            &\label{eq:p1_3}\leq \frac{1}{N}\sum_{n=1}^{N} \max_{c \in [C]} \left \{ \left (\mathbf{v}_{(c)} \right )^{\mathsf{T}}\mathbf{M}\widehat{\mathbf{x}}_{(n)} - \widehat{\mathbf{y}}_{(n)}^{\mathsf{T}}\mathbf{M}\widehat{\mathbf{x}}_{(n)} - \left (\mathbf{v}_{(c)} \right )^{\mathsf{T}}\widehat{\mathbf{y}}_{(n)} + 1 \right \}\\
            &\label{eq:p1_4}= \frac{1}{N}\sum_{n=1}^{N} \max_{c \in [C]} \left \{ \left ( \mathbf{v}_{(c)} \right )^\mathsf{T} \left ( \mathbf{M} \mathbf{x} - \mathbf{y} \right ) + 1 \right \} - \mathbf{y}^{\mathsf{T}} \mathbf{M} \mathbf{x}\\
            &\label{eq:p1_5}= \frac{1}{N}\sum_{n=1}^{N} \ell_{\text{CS}}(\mathbf{M};\widehat{\boldsymbol{\xi}}_{(n)}),
        \end{align}
    \end{subequations}
    where \eqref{eq:p1_1} follows by noting that the empirical error is $0$ if the classification is correct and $1$ otherwise, however the expression on the right-hand side is $0$ if the classification is correct but greater than or equal to $1$ otherwise. The inequality \eqref{eq:p1_3} is obtained by noting that $H_{\mathbf{M}}(\widehat{\mathbf{x}}_{(n}) \in \{ \mathbf{v}_{(c)} \}_{c=1}^C$, and finally \eqref{eq:p1_4} is obtained by grouping the terms.

    \item We first derive an upper bounding function $f(\boldsymbol{\xi})$ for $ \ell_{\text{CS}} (\mathbf{M};\boldsymbol{\xi}) $ as follows:
    \begin{subequations}
    \begin{align}
        \ell_{\text{CS}} (\mathbf{M};\boldsymbol{\xi}) = \ell_{\mathbf{M}} \left ( (\mathbf{x};\mathbf{y}) \right ) &\label{eq:p2_1_1}= \max_{c \in [C]} \left \{ \left ( \mathbf{v}_{(c)} \right )^\mathsf{T} \left ( \mathbf{M} \mathbf{x} - \mathbf{y} \right ) + 1 \right \} - \mathbf{y}^{\mathsf{T}} \mathbf{M} \mathbf{x} \\
        &\label{eq:p2_1_2} \leq \max_{c \in [C]} \left \{ \left ( \mathbf{v}_{(c)} \right )^\mathsf{T} \mathbf{M} \mathbf{x} + 1 \right \} - \mathbf{y}^{\mathsf{T}} \mathbf{M} \mathbf{x} \\
        &\label{eq:p2_1_3} \leq \max_{c \in [C]} \left \{ \left ( \mathbf{v}_{(c)} \right )^\mathsf{T} \mathbf{M} \mathbf{x} + 1 \right \} -  \min_{j \in [C]} \left \{ \left ( \mathbf{v}_{(j)} \right )^\mathsf{T} \mathbf{M} \mathbf{x} \right \}\\
        &\label{eq:p2_1_4}   \coloneqq f(\boldsymbol{\xi}) \nonumber,
    \end{align}
    \end{subequations}
    where \eqref{eq:p2_1_1} follows from recalling that all the elements of $\mathbf{v}_{(c)}$ and $\mathbf{y}$ are non-negative, and \eqref{eq:p2_1_3} follows by noting that $\mathbf{y} \in \left \{ \mathbf{v}_{(j)} \right \}_{j=1}^C$. Now, we analyze the Lipschitz continuity of $f(\boldsymbol{\xi})$ as follows:
    \begin{subequations}
    \begin{align}
    \left | f(\boldsymbol{\xi}_{(1)}) - f(\boldsymbol{\xi}_{(2)}) \right | & = \left | \max_{c_1 \in [C]} \left \{ \left ( \mathbf{v}_{(c_1)} \right )^\mathsf{T} \mathbf{M} \mathbf{x}_{(1)} + 1 \right \} -  \min_{j_1 \in [C]} \left \{ \left ( \mathbf{v}_{(j_1)} \right )^\mathsf{T} \mathbf{M} \mathbf{x}_{(1)} \right \} \right. \nonumber \\
    & \qquad \qquad \left. - \max_{c_2 \in [C]} \left \{ \left ( \mathbf{v}_{(c_2)} \right )^\mathsf{T} \mathbf{M} \mathbf{x}_{(2)} + 1 \right \} +  \min_{j_2 \in [C]} \left \{ \left ( \mathbf{v}_{(j_2)} \right )^\mathsf{T} \mathbf{M} \mathbf{x}_{(2)} \right \} \right |\\
    &\label{eq:p2_2_2} \leq \left | \max_{c_1 \in [C]} \left \{ \left ( \mathbf{v}_{(c_1)} \right )^\mathsf{T} \mathbf{M} \mathbf{x}_{(1)} \right \} -  \max_{c_2 \in [C]} \left \{ \left ( \mathbf{v}_{(c_2)} \right )^\mathsf{T} \mathbf{M} \mathbf{x}_{(2)} \right \}\right | \nonumber \\
    & \qquad \qquad \qquad + \left | \min_{j_1 \in [C]} \left \{ \left ( \mathbf{v}_{(j_1)} \right )^\mathsf{T} \mathbf{M} \mathbf{x}_{(1)} \right \} - \min_{j_2 \in [C]} \left \{ \left ( \mathbf{v}_{(j_2)} \right )^\mathsf{T} \mathbf{M} \mathbf{x}_{(2)} \right \} \right | \\
    & \label{upper_bd_diff} = \left | \left ( \mathbf{v}_{(c^{\ast}_1)} \right )^\mathsf{T} \mathbf{M} \mathbf{x}_{(1)} - \left ( \mathbf{v}_{(c^{\ast}_2)} \right )^\mathsf{T} \mathbf{M} \mathbf{x}_{(2)} \right | + \nonumber \\
    & \qquad \qquad \qquad \left | \left ( \mathbf{v}_{(j^{\ast}_1)} \right )^\mathsf{T} \mathbf{M} \mathbf{x}_{(1)} - \left ( \mathbf{v}_{(j^{\ast}_2)} \right )^\mathsf{T} \mathbf{M} \mathbf{x}_{(2)} \right |,
    \end{align}
    \end{subequations}
    where \eqref{eq:p2_2_2} follows from the triangle inequality, and to get \eqref{upper_bd_diff} we assume, without loss of generality, that $\mathbf{v}_{(c^{\ast}_1)}$, $\mathbf{v}_{(c^{\ast}_2)}$, $\mathbf{v}_{(j^{\ast}_1)}$, and $\mathbf{v}_{(j^{\ast}_2)}$ are the optimizers of their respective maximization and minimization problems. We start by analyzing the first term, where we consider three cases as follows:
    \begin{enumerate}
        \item \label{item:equal_case} $\mathbf{v}_{(c^{\ast}_1)} = \mathbf{v}_{(c^{\ast}_2)}$. In this case we get the following:
            \begin{subequations}
            \begin{align}
                \left | \left ( \mathbf{v}_{(c^{\ast}_1)} \right )^\mathsf{T} \mathbf{M} \mathbf{x}_{(1)} - 
            \left ( \mathbf{v}_{(c^{\ast}_2)} \right )^\mathsf{T} \mathbf{M} \mathbf{x}_{(2)} \right | &\label{eq:p2_3_1} = \left | \left ( \mathbf{v}_{(c^{\ast}_1)} \right )^\mathsf{T} \mathbf{M} \mathbf{x}_{(1)} - 
            \left ( \mathbf{v}_{(c^{\ast}_1)} \right )^\mathsf{T} \mathbf{M} \mathbf{x}_{(2)} \right |\\
            &\label{eq:p2_3_2}= \left | \left ( \mathbf{v}_{(c^{\ast}_1)} \right )^\mathsf{T}\mathbf{M}(\mathbf{x}_{(1)} - \mathbf{x}_{(2)}) \right |\\
            &\label{eq:p2_3_3}\leq ||\mathbf{v}_{(c^{\ast}_1)}||_{\ast} ||\mathbf{M}(\mathbf{x}_{(1)} - \mathbf{x}_{(2)}) ||\\
            &\label{eq:p2_3_4}= ||\mathbf{M}(\mathbf{x}_{(1)} - \mathbf{x}_{(2)}) ||\\
            &\label{eq:p2_3_5} \leq \left | \left | \mathbf{M} \right | \right | \left | \left | \mathbf{x}_{(1)} - \mathbf{x}_{(2)} \right | \right |\\
            &\label{eq:p2_3_6} \leq \left | \left | \mathbf{M} \right | \right | c(\boldsymbol{\xi}_{(1)},\boldsymbol{\xi}_{(2)}),
            \end{align}
            \end{subequations}
       where \eqref{eq:p2_3_3} follows from the Cauchy-Schwarz inequality, \eqref{eq:p2_3_4} follows from the fact that $\mathbf{v}_{(c^{\ast}_1)}$ is a unit vector, and $||\mathbf{M}||$ is the matrix norm of $\mathbf{M}$ induced by the corresponding vector norm $||\cdot||$.
        \item  $\mathbf{v}_{(c^{\ast}_1)} \neq \mathbf{v}_{(c^{\ast}_2)}$ and $\left ( \mathbf{v}_{(c^{\ast}_1)} \right )^\mathsf{T} \mathbf{M} \mathbf{x}_{(1)} \geq 
        \left ( \mathbf{v}_{(c^{\ast}_2)} \right )^\mathsf{T} \mathbf{M} \mathbf{x}_{(2)}$. In this case, we note that:
        \begin{subequations}
        \begin{align}
        \left | \left ( \mathbf{v}_{(c^{\ast}_1)} \right )^\mathsf{T} \mathbf{M} \mathbf{x}_{(1)} - \left ( \mathbf{v}_{(c^{\ast}_2)} \right )^\mathsf{T} \mathbf{M} \mathbf{x}_{(2)} \right | &\label{eq:p2_4_1} \leq \left | \left ( \mathbf{v}_{(c^{\ast}_1)} \right )^\mathsf{T} \mathbf{M} \mathbf{x}_{(1)} - 
        \left ( \mathbf{v}_{(c^{\ast}_1)} \right )^\mathsf{T} \mathbf{M} \mathbf{x}_{(2)} \right |\\
        &\label{eq:p2_4_2} \leq \left | \left | \mathbf{M} \right | \right | c(\boldsymbol{\xi}_{(1)},\boldsymbol{\xi}_{(2)}),
        \end{align}
        \end{subequations}
        where \eqref{eq:p2_4_1} results from noting that $\mathbf{v}_{(c^{\ast}_1)}$ and $\mathbf{v}_{(c^{\ast}_2)}$ are the maximizers of their respective maximization problems, and \eqref{eq:p2_4_2} is obtained by following the same logic used in the previous Case \ref{item:equal_case}.
        \item $\mathbf{v}_{(c^{\ast}_1)} \neq \mathbf{v}_{(c^{\ast}_2)}$ and $\left ( \mathbf{v}_{(c^{\ast}_1)} \right )^\mathsf{T} \mathbf{M} \mathbf{x}_{(1)} < 
        \left ( \mathbf{v}_{(c^{\ast}_2)} \right )^\mathsf{T} \mathbf{M} \mathbf{x}_{(2)}$. In this case, we note that:
        \begin{subequations}
        \begin{align}
        \left | \left ( \mathbf{v}_{(c^{\ast}_1)} \right )^\mathsf{T} \mathbf{M} \mathbf{x}_{(1)} - \left ( \mathbf{v}_{(c^{\ast}_2)} \right )^\mathsf{T} \mathbf{M} \mathbf{x}_{(2)} \right | & \leq \left | \left ( \mathbf{v}_{(c^{\ast}_2)} \right )^\mathsf{T} \mathbf{M} \mathbf{x}_{(1)} - 
        \left ( \mathbf{v}_{(c^{\ast}_2)} \right )^\mathsf{T} \mathbf{M} \mathbf{x}_{(2)} \right |\\
        & \leq \left | \left | \mathbf{M} \right | \right | c(\boldsymbol{\xi}_{(1)},\boldsymbol{\xi}_{(2)}),
        \end{align}
        \end{subequations}
        where we use the same logic used in the previous two cases.
    \end{enumerate}
    Similarly, when we consider the second term in \eqref{upper_bd_diff}, it also suffices to study three different cases in the same manner we studied the first term. Doing this results in the following:
    \begin{equation}
        \left | \left ( \mathbf{v}_{(j^{\ast}_1)} \right )^\mathsf{T} \mathbf{M} \mathbf{x}_{(1)} - \left ( \mathbf{v}_{(j^{\ast}_2)} \right )^\mathsf{T} \mathbf{M} \mathbf{x}_{(2)} \right | \leq \left | \left | \mathbf{M} \right | \right | c(\boldsymbol{\xi}_{(1)},\boldsymbol{\xi}_{(2)}).
    \end{equation}
    Therefore, we can see that:
    \begin{subequations}
    \begin{align}
        \left | f(\boldsymbol{\xi}_{(1)}) - f(\boldsymbol{\xi}_{(2)}) \right | & \leq \left | \left ( \mathbf{v}_{(c^{\ast}_1)} \right )^\mathsf{T} \mathbf{M} \mathbf{x}_{(1)} - \left ( \mathbf{v}_{(c^{\ast}_2)} \right )^\mathsf{T} \mathbf{M} \mathbf{x}_{(2)} \right | + \nonumber \\ & \qquad \qquad \qquad \left | \left ( \mathbf{v}_{(j^{\ast}_1)} \right )^\mathsf{T} \mathbf{M} \mathbf{x}_{(1)} - \left ( \mathbf{v}_{(j^{\ast}_2)} \right )^\mathsf{T} \mathbf{M} \mathbf{x}_{(2)} \right |\\
        & \leq 2\left | \left | \mathbf{M} \right | \right | c(\boldsymbol{\xi}_{(1)},\boldsymbol{\xi}_{(2)}),
    \end{align}
    \end{subequations}
    indicating that $f(\boldsymbol{\xi})$ is Lipschitz continuous with $\Lip \left ( f(\boldsymbol{\xi}) \right ) \leq 2\left | \left | \mathbf{M} \right | \right |$.

    \item This follows from the fact that $\ell_{\text{CS}}(\mathbf{M};\boldsymbol{\xi})$ is a maximum of convex functions in $\mathbf{x}$.

    \item This follows from the fact that $\ell_{\text{CS}}(\mathbf{M};\boldsymbol{\xi})$ is a maximum of convex functions in $\mathbf{M}$.

    \item For this part, it suffices to study the Lipschitz continuity of the function $\ell_{\text{CS},c}(\mathbf{M};(\mathbf{x},\widecheck{\mathbf{y}})) = \left ( \mathbf{v}_{(c)}^{\mathsf{T}} \right ) \mathbf{M} \mathbf{x} - \widecheck{\mathbf{y}}^{\mathsf{T}}\mathbf{M}\mathbf{x}$ for a fixed $\widecheck{\mathbf{y}} \in \mathcal{Y}$ since the discarded terms do not depend on $\mathbf{x}$. We do that as follows
    \begin{subequations}
    \begin{align}
        \Lip(\ell_{\text{CS},c}(\mathbf{M};(\mathbf{x},\widecheck{\mathbf{y}}))) &\label{eq:p5_1}= \sup_{\boldsymbol{\omega}} \left \{ \left | \left | \boldsymbol{\omega} \right | \right |_{\ast} \colon \ell_{\text{CS},c}^{\ast}(\mathbf{\boldsymbol{\omega}}) < \infty \right \}\\
        &\label{eq:p5_2} = \sup_{\boldsymbol{\omega}} \left \{ \left | \left | \boldsymbol{\omega} \right | \right |_{\ast} \colon \sup_{\mathbf{x}} \left \{ \boldsymbol{\omega}^{\mathsf{T}}\mathbf{x} - \left ( \mathbf{v}_{(c)} - \widecheck{\mathbf{y}}\right )^{\mathsf{T}}\mathbf{M} \mathbf{x}  \right \} < \infty \right \}\\
        &\label{eq:p5_3} = \sup_{\boldsymbol{\omega}} \left \{ \left | \left | \boldsymbol{\omega} \right | \right |_{\ast} \colon \boldsymbol{\omega} = \left ( \mathbf{v}_{(c)} - \widecheck{\mathbf{y}}\right )^{\mathsf{T}}\mathbf{M} \right \}\\
        &\label{eq:p5_4} = \left | \left | \left ( \mathbf{v}_{(c)} - \widecheck{\mathbf{y}} \right )^{\mathsf{T}} \mathbf{M}\right | \right |_{\ast},
    \end{align}
    \end{subequations}
    where \eqref{eq:p5_1} follows from the definition of the Lipschitz modulus, \eqref{eq:p5_2} follows from the definition of the conjugate $\ell_{\text{CS},c}^{\ast}(\mathbf{\boldsymbol{\omega}})$ of the constituent $\ell_{\text{CS},c}(\mathbf{M};(\mathbf{x},\widecheck{\mathbf{y}}))$, and \eqref{eq:p5_3} is obtained by noting that this is the only condition under which the inner maximization problem is bounded.
\end{enumerate}
\subsection{Proof of Theorem \ref{thm:linear_DRMSVM}}
\begin{proof}
    In this proof we follow very similar strategies to the ones used by \cite{2019regularization,datadrivenDRO,shafieezadehabadeh2015distributionally}. More specifically, we begin by rewriting the inner risk maximization problem from the DRO problem in \eqref{eq:dro_problem} as follows:
    \begin{subequations}
    \begin{align}
    &\sup_{\mathbb{Q} \in \mathcal{A}_{\varepsilon,1}} \mathbb{E}^{\mathbb{Q}} \left [ \ell_{\text{CS}} (\mathbf{M};\boldsymbol{\xi}) \right ] \nonumber \\&\label{eq:t1_1_0}= \inf_{\lambda \geq 0} \lambda \varepsilon + \frac{1}{N} \sum_{n=1}^{N} \sup_{\boldsymbol{\xi} \in \Xi} \ell_{\text{CS}} (\mathbf{M};\boldsymbol{\xi}) - \lambda c(\boldsymbol{\xi},\widehat{\boldsymbol{\xi}}_{(n)})\\
     &\label{eq:t1_1_1}= \left \{ \begin{aligned}
        &\inf_{\lambda,s_n} && \lambda \varepsilon + \frac{1}{N} \sum_{n=1}^{N} s_n &&\\
        & \subjectto && \sup_{\boldsymbol{\xi} \in \Xi} \left \{ \ell_{\text{CS}} (\mathbf{M};\boldsymbol{\xi}) - \lambda c(\boldsymbol{\xi},\widehat{\boldsymbol{\xi}}_{(n)}) \right \} \leq s_n && \forall n \in [N]\\
        &&& \lambda \geq 0
     \end{aligned} \right.\\
     &\label{eq:t1_1_2}= \left \{ \begin{aligned}
        &\inf_{\lambda,s_n} && \lambda \varepsilon + \frac{1}{N} \sum_{n=1}^{N} s_n&&\\
        & \subjectto && \sup_{\mathbf{x} \in \mathcal{X}, \mathbf{y} \in \mathcal{Y}} \{ \ell_{\text{CS}} (\mathbf{M};(\mathbf{x},\mathbf{y})) - \lambda c(\boldsymbol{\xi},\widehat{\boldsymbol{\xi}}_{(n)}) \} \leq s_n && \forall n \in [N]\\
        &&& \lambda \geq 0
     \end{aligned} \right.\\
     &\label{eq:t1_1_3} = \left \{ \begin{aligned}
        &\inf_{\lambda,s_n} && \lambda \varepsilon + \frac{1}{N} \sum_{n=1}^{N} s_n&&\\
        & \subjectto && \sup_{\mathbf{x} \in \mathcal{X}} \left \{ \ell_{\text{CS}} (\mathbf{M};(\mathbf{x},\widehat{\mathbf{y}}_{(n)})) - \lambda \left | \left | \mathbf{x} - \widehat{\mathbf{x}}_{(n)} \right | \right | \right \} \leq s_n && \forall n \in [N]\\
        &&& \sup_{\mathbf{x} \in \mathcal{X}} \left \{ \ell_{\text{CS}} (\mathbf{M};(\mathbf{x},\widecheck{\mathbf{y}}_{(c)})) - \lambda \left | \left | \mathbf{x} - \widehat{\mathbf{x}}_{(n)} \right | \right | \right \} - \lambda \kappa \leq s_n && \begin{array}{l} \forall n \in [N] \\ \forall \widecheck{\mathbf{y}}_{(c)} \in \mathcal{Y}, \\ \widecheck{\mathbf{y}}_c \neq \widehat{\mathbf{y}}_n \end{array} \\
        &&& \lambda \geq 0
     \end{aligned}\right.\\
     & \label{near_final_prog2} = \left \{ \begin{aligned}
         &\inf_{\lambda,s_n} && \lambda \varepsilon + \frac{1}{N} \sum_{n=1}^{N} s_n\\
        & \subjectto &&\begin{multlined} \sup_{\mathbf{x} \in \mathbb{R}^P} \left \{ \left (\mathbf{v}_{(j)} - \widehat{\mathbf{y}}_{(n)}\right )^{\mathsf{T}}\mathbf{M}\mathbf{x} - \mathbf{v}_{(j)}^{\mathsf{T}}\widehat{\mathbf{y}}_{(n)} - \lambda \left | \left | \mathbf{x} - \widehat{\mathbf{x}}_{(n)} \right | \right | \right \} \leq s_n \\ \forall n \in [N], \ \forall j \in [C] \end{multlined}\\
        &&& \begin{multlined} \sup_{\mathbf{x} \in \mathbb{R}^P} \left \{ \left (\mathbf{v}_{(j)} - \widecheck{\mathbf{y}}_{(c)}\right )^{\mathsf{T}}\mathbf{M}\mathbf{x} - \mathbf{v}_{(j)}^{\mathsf{T}}\widecheck{\mathbf{y}}_{(c)} - \lambda \left | \left | \mathbf{x} - \widehat{\mathbf{x}}_{(n)} \right | \right | \right \} - \lambda \kappa \leq s_n \\
        \forall n \in [N] \ \forall j \in [C] \ \forall \widecheck{\mathbf{y}}_{(c)} \in \mathcal{Y}, \widecheck{\mathbf{y}}_c \neq \widehat{\mathbf{y}}_n \end{multlined}&&\\
        &&& \lambda \geq 0
     \end{aligned}\right.
    \end{align}\end{subequations}
    where \eqref{eq:t1_1_0} follows from Lemma A.1 in \cite{2019regularization} by noting Lemma \ref{lem:prop_loss} Part \ref{part:bounded_above}, \eqref{eq:t1_1_1} follows by moving the inner maximization problem to the constraints and introducing slack variables $s_n$ for all $n \in [N]$. The equality in \eqref{eq:t1_1_3} follows by observing that $y$ can take a finite number of discrete values, thus we can leverage that fact to remove it from the maximization problem and consider each possible value of $\mathbf{y}$ individually. Finally, \eqref{near_final_prog2} follows by observing that $\mathbf{v}_{(j)}$ too can only take a finite number of discrete values. Thus, we eliminate the maximization in the definition of $\ell_{\text{CS}}(\mathbf{M};\boldsymbol{\xi})$, and instead impose the constraint that the slack variables $s_n$ are greater than or equal to all the individual loss values obtained by $\mathbf{v}_{(j)}$ for $j \in [C]$. We also replace the support set $\mathcal{X}$ of the features with $\mathbb{R}^P$ a stated in Asm. \ref{assump:feature_support}. To simplify the notation, let us introduce the following
    \begin{equation}
        \widecheck{\ell}_{\text{CS}}(\mathbf{x}) = (\mathbf{v}_{(j)} - \widecheck{\mathbf{y}}_{(c)})^{\mathsf{T}}\mathbf{M}\mathbf{x} - \mathbf{v}_{(j)}^{\mathsf{T}}\widecheck{\mathbf{y}}_{(c)},
    \end{equation}
    where $j \in [C]$ and $\widecheck{\mathbf{y}}_{(c)} \in \mathcal{Y}$. It follows from Lemma \ref{lem:prop_loss} that $\widecheck{\ell}_{\text{CS}}(\mathbf{x})$ is convex and Lipschitz continuous in $\mathbf{x}$. Therefore, $\widecheck{\ell}_{\text{CS}}(\mathbf{x})$ coincides with its bi-conjugate function $\widecheck{\ell}_{\text{CS}}^{\ast \ast}(\mathbf{x})$. Therefore, we can rewrite it as
    \begin{equation}
        \widecheck{\ell}_{\text{CS}}(\mathbf{x}) = \sup_{\boldsymbol{\theta} \in \Theta} \left \{ \langle \boldsymbol{\theta},\mathbf{x} \rangle - \widecheck{\ell}^{\ast}_{\text{CS}}(\boldsymbol{\theta}) \right \},
    \end{equation}
    where $\Theta \coloneqq \{\boldsymbol{\theta} \in \mathbb{R}^P :\widecheck{\ell}^{\ast}_{\text{CS}}(\boldsymbol{\theta}) < \infty \} $ is the effective domain of the conjugate function $\widecheck{\ell}^{\ast}_{\text{CS}}(\boldsymbol{\theta})$. Using this representation of the loss function along with the definition of the dual norm enables us to write the following:
    \begin{subequations}
    \begin{align}
        \sup_{\mathbf{x} \in \mathbb{R}^P} \left \{ \widecheck{\ell}_{\text{CS}}(\mathbf{x}) - \lambda \left | \left | \mathbf{x} - \widehat{\mathbf{x}}_{(n)} \right | \right | \right \} & = \sup_{\mathbf{x} \in \mathbb{R}^P} \sup_{\boldsymbol{\theta} \in \Theta} \left \{\left \langle \theta,\mathbf{x} \right \rangle - \widecheck{\ell}^{\ast}_{\text{CS}}(\boldsymbol{\theta}) - \lambda \left | \left | \mathbf{x} - \widehat{\mathbf{x}}_{(n)} \right | \right | \right \}\\
        & = \sup_{\mathbf{x} \in \mathbb{R}^P} \sup_{\boldsymbol{\theta} \in \Theta} \inf_{\left | \left | \mathbf{a} \right | \right |_{\ast} \leq \lambda} \left \{ \left \langle \boldsymbol{\theta},\mathbf{x} \right \rangle - \widecheck{\ell}^{\ast}_{\text{CS}}(\boldsymbol{\theta}) + \left \langle \mathbf{a}, \mathbf{x} \right \rangle - \left \langle \mathbf{a}, \widehat{\mathbf{x}}_{(n)} \right \rangle \right \}.
    \end{align}
    \end{subequations}
    We then utilize Proposition 5.5.4 in \cite{bertsekas2009convex} to swap the maximization over $\mathbf{x}$ with the maximization over $\boldsymbol{\theta}$, and then the minimization over $\mathbf{a}$. This allows us to obtain the following:
    \begin{subequations}
    \begin{align}
        \sup_{\mathbf{x} \in \mathbb{R}^P} \{ \widecheck{\ell}_{\text{CS}}(\mathbf{x}) - \lambda \left | \left | \mathbf{x} - \widehat{\mathbf{x}}_{(n)} \right | \right | \} & = \sup_{\boldsymbol{\theta} \in \Theta} \inf_{\left | \left | \mathbf{a} \right | \right |_{\ast} \leq \lambda} \sup_{\mathbf{x} \in \mathbb{R}^P} \left \{ \left \langle \boldsymbol{\theta} + \mathbf{a}, \mathbf{x} \right \rangle - \widecheck{\ell}^{\ast}_{\text{CS}}(\boldsymbol{\theta}) - \left \langle \mathbf{a},\widehat{\mathbf{x}}_{(n)} \right \rangle \right \}\\
        & \label{eq:sup_inf_prob} = \sup_{\boldsymbol{\theta} \in \Theta} \inf_{\left | \left | \mathbf{a} \right | \right |_{\ast} \leq \lambda} \left \{ \sigma_{\mathbb{R}^P}(\theta + \mathbf{a})- \widecheck{\ell}^{\ast}_{\text{CS}}(\boldsymbol{\theta}) - \left \langle \mathbf{a},\widehat{\mathbf{x}}_{(n)} \right \rangle \right \},
    \end{align}
    \end{subequations}
    where $\sigma_{\mathbb{R}^P}(\mathbf{z}) = \sup_{\mathbf{x} \in \mathbb{R}^P}\langle \mathbf{x}, \mathbf{z} \rangle$ is the support function of $\mathbb{R}^P$. We can leverage the fact that $\sigma_{\mathbb{R}^P}(\boldsymbol{\theta} + \mathbf{a}) = \chi_{\{ 0 \}}(\boldsymbol{\theta} + \mathbf{a})$, where $\chi_{\mathcal{S}}$ is the characteristic function of the set $\mathcal{S}$, defined as
    \begin{equation}
        \chi_{\mathcal{S}}(\mathbf{z}) = \left \{ \begin{aligned}
            & 0  &&\text{if } \mathbf{z} \in \mathcal{S},\\
            & +\infty &&\text{otherwise.}
        \end{aligned} \right..
    \end{equation}
    Using this, we rewrite \eqref{eq:sup_inf_prob} as follows
    \begin{subequations}
    \begin{align}
        \sup_{\mathbf{x} \in \mathbb{R}^P} \left \{ \widecheck{\ell}_{\text{CS}}(\mathbf{x}) - \lambda \left | \left | \mathbf{x} - \widehat{\mathbf{x}}_{(n)} \right | \right | \right \} &\label{eq:t1_2_0}= \sup_{\boldsymbol{\theta} \in \Theta} \inf_{\left | \left | \mathbf{a} \right | \right |_{\ast} \leq \lambda} \left \{ \chi_{\{0\}}(\boldsymbol{\theta} + \mathbf{a}) - \widecheck{\ell}^{\ast}_{\text{CS}}(\boldsymbol{\theta}) - \left \langle \mathbf{a},\widehat{\mathbf{x}}_{(n)} \right \rangle \right \} \\
        &\label{eq:t1_2_1} = \left \{ \begin{aligned}
            & \sup_{\boldsymbol{\theta} \in \Theta} \inf_{\left | \left | \mathbf{a} \right | \right |_{\ast} \leq \lambda} \left \{  - \widecheck{\ell}^{\ast}_{\text{CS}}(\boldsymbol{\theta}) - \left \langle \mathbf{a},\widehat{\mathbf{x}}_{(n)} \right \rangle \right \} && \text{if } \boldsymbol{\theta} = -\mathbf{a},\\
            & +\infty && \text{otherwise}
        \end{aligned} \right.\\
        &\label{eq:t1_2_2} \leq \left \{ \begin{aligned}
            & \sup_{\boldsymbol{\theta} \in \Theta} \left \{ \left \langle \theta,\widehat{\mathbf{x}}_n \right \rangle - \widecheck{\ell}^{\ast}_{\text{CS}}(\boldsymbol{\theta}) \right \} && \text{if} \ \sup \left \{ \left | \left | \boldsymbol{\theta} \right | \right |_{\ast} \colon \boldsymbol{\theta} \in \Theta \right \} \leq \lambda,\\
            & +\infty && \text{otherwise.}
        \end{aligned} \right.\\
        &\label{eq:t1_2_3} = \left \{ \begin{aligned}
            & \widecheck{\ell}_{\text{CS}}(\widehat{\mathbf{x}}_{(n)}) && \text{if} \ \sup \left \{ \left | \left | \theta \right | \right |_s \colon \theta \in \Theta \right \} \leq \lambda,\\
            & \infty && \text{otherwise.}
        \end{aligned}\right.\\
        & \label{sup_up_bd} = \left \{ \begin{aligned}
            & \widecheck{\ell}_{\text{CS}}(\widehat{\mathbf{x}}_{(n)}) && \text{if} \ \Lip \left (\widecheck{\ell}_{\text{CS}}(\mathbf{x}) \right ) \leq \lambda,\\
            & \infty && \text{otherwise.}
        \end{aligned}\right.\\
        &\label{eq:t1_2_5} = \left \{ \begin{aligned}
            & \widecheck{\ell}_{\text{CS}}(\widehat{\mathbf{x}}_{(n)}) && \text{if} \ \left | \left | \left ( \mathbf{v}_{(j)}-\widecheck{\mathbf{y}}_{(c)} \right )^{\mathsf{T}}\mathbf{M} \right | \right |_{\ast} \leq \lambda,\\
            & \infty && \text{otherwise.}
        \end{aligned} \right.
    \end{align}
    \end{subequations}
    where \eqref{eq:t1_2_1} follows from the definition of $\chi_{\{ 0 \}}$, the inequality in \eqref{eq:t1_2_2} follows from substituting $\mathbf{a}$ with $-\boldsymbol{\theta}$, and \eqref{eq:t1_2_3} follows from the fact that $\widecheck{\ell}_{\text{CS}}(\mathbf{x})$ coincides with its bi-conjugate function. The equality in \eqref{sup_up_bd} follows from the definition of the Lipschitz modulus, and finally \eqref{eq:t1_2_5} follows from Lemma \ref{lem:prop_loss} Part \ref{part:lip_mod}. We note that the condition in the final expression must hold for all the constraints in the problem, thus it must hold for all $j \in [C]$ and all $\widecheck{\mathbf{y}}_{(c)} \in \mathcal{Y}$. Thus, we plug this result into the program and obtain our final result by including $\mathbf{M}$ as a decision variable.
\end{proof}
\subsection{Proof of Theorem \ref{thm:kernel_DRMSVM}}
\begin{proof}
This proof follows a very similar strategy to that used in \cite{2019regularization} to derive a kernel version of the DR binary SVM. Suppose that the separations between different classes are hypotheses $\mathbf{h}$ that belong to a reproducing kernel Hilbert space (RKHS) $\mathbb{H} \subseteq \mathbb{R}^{\mathcal{X}}$. This space is equipped with a self-dual norm $||\cdot||_{\mathbb{H}}$ induced by the inner product $\langle \cdot, \cdot \rangle_{\mathbb{H}}$.

\begin{theorem}[Riesz Representation Theorem \cite{conway2007}] Let $\mathscr{F}: \mathbb{H} \rightarrow \mathbb{R}$ be a continuous linear functional, then for every $h \in \mathbb{H}$ there exists a unique $\mathbf{h}_{(0)} \in \mathbb{H}$ such that $\mathscr{F}(\mathbf{h}) = \langle \mathbf{h},\mathbf{h}_{(0)} \rangle_{\mathbb{H}}$.
\end{theorem}

Now, consider the sampling functional $\mathscr{F}_{\text{s}}(\mathbf{h}) = \mathbf{h}(\mathbf{x}) \ \forall \mathbf{h} \in \mathbb{H}$. This functional is linear. Moreover, it is continuous in many infinite dimensional RKHSs. Thus, it is susceptible to the Riesz representation theorem. We introduce a feature map $\Psi:\mathcal{X} \rightarrow \mathbb{H}$ such that $\mathbf{h}(\mathbf{x}) = \langle \mathbf{h}, \Psi(\mathbf{x}) \rangle_{\mathbb{H}} \  \forall \mathbf{x} \in \mathcal{X}$. This feature map gives rise to a kernel function $k:\mathcal{X} \times \mathcal{X} \rightarrow \mathbb{R}_+$, where $k(\mathbf{x}_{(1)},\mathbf{x}_{(2)}) = \langle \Psi (\mathbf{x}_{(1)}),\Psi (\mathbf{x}_{(2)}) \rangle_{\mathbb{H}}$. As illustrated in \cite{2019regularization}, the function $k$ is symmetric and positive semi-definite (PSD) by construction. Therefore, by Moore-Aronszajn Theorem it possesses the reproducing property (i.e. it uniquely defines an RKHS) \cite{Moore1935,Aron1950}. 

Now, suppose we have a set of hypotheses $\{\mathbf{h}_{(c)}\}_{c=1}^{C} \subset \mathbb{H}$ representative of the nonlinear separations between classes. We define $\mathbf{H} \coloneqq \left [\mathbf{h}_{(1)}^{\mathsf{T}};\dots;\mathbf{h}_{(C)}^{\mathsf{T}} \right ]$. Consequently, we have
\begin{equation*}
    \mathbf{H}(\mathbf{x}) =  \sum_{n=1}^{\infty} \boldsymbol{\alpha}_{(n)}k(\mathbf{x}_{(n)},\mathbf{x}),
\end{equation*}
where $\boldsymbol{\alpha}_{(n)} \in \mathbb{R}^C$ is such that the entry $\boldsymbol{\alpha}_{(n)_i}$ is the coefficient $\alpha_n$ for $\mathbf{h}_{(i)}$. Next, suppose that the input features $\mathbf{x} \in \mathcal{X}$ can be replaced with features $\mathbf{x}_{(\mathbb{H})} \in \mathbb{H}$. Similarly, suppose that each nonlinear hypothesis $\mathbf{h}_{(c)} \in \mathbb{H}$ is identified by a linear hypothesis $\mathbf{h}_{(c)_{(\mathbb{H})}}(\mathbf{x}_{(\mathbb{H})}) \in \mathbb{H}$ via $\mathbf{h}_{(c)_{(\mathbb{H})}}(\mathbf{x}_{(\mathbb{H})}) = \langle \mathbf{h}_{(c)},\mathbf{x}_{(\mathbb{H})} \rangle_{\mathbb{H}} =  \langle \mathbf{h}_{(c)},\Psi(\mathbf{x}) \rangle_{\mathbb{H}} = \mathbf{h}_{(c)}(\mathbf{x})$. To simplify the notation, we introduce $\mathbf{g}_{(\mathbb{H})}(\mathbf{x}_{(\mathbb{H})}) \coloneqq \left [\mathbf{h}_{(1)_{(\mathbb{H})}}(\mathbf{x}_{(\mathbb{H})})^{\mathsf{T}};\dots;\mathbf{h}_{(C)_{(\mathbb{H})}}(\mathbf{x}_{(\mathbb{H})})^{\mathsf{T}} \right ]$. This allows us to rewrite our loss function as
\begin{equation*}
    \ell_{\text{CS},\mathbb{H}}(\mathbf{g}_{(\mathbb{H})}(\mathbf{x}_{(\mathbb{H})});\mathbf{y}) = \max_{c \in [C]} \left \{ \mathbf{v}_{(c)}^{\mathsf{T}}\left (\mathbf{g}_{(\mathbb{H})}(\mathbf{x}_{(\mathbb{H})}) - \mathbf{y} \right ) + 1 \right \} - \mathbf{y}^{\mathsf{T}}\mathbf{g}_{(\mathbb{H})}(\mathbf{x}_{(\mathbb{H})}).
\end{equation*}
This in turn allows us to write a lifted version of the DRO problem in \eqref{eq:dro_problem} as
\begin{equation}\label{eq:lifted_nonlin}
    \inf_{\{ \mathbf{h}_{(c)}\}_{c=1}^C \in \mathbb{H}} \sup_{\mathbb{Q} \in \mathcal{A}_{\varepsilon,1}^{\mathbb{H}}(\Xi)} \mathbb{E}^{\mathbb{Q}} [\ell_{\text{CS},\mathbb{H}}(\mathbf{g}_{(\mathbb{H})}(\mathbf{x}_{(\mathbb{H})});\mathbf{y})],
\end{equation}
where $\mathcal{A}_{\varepsilon,1}^{\mathbb{H}}(\Xi)$ is a type-1 Wasserstein ball with radius $\varepsilon$ centered at $\widehat{\mathbb{P}}_N^{\mathbb{H}} = \frac{1}{N}\sum_{n=1}^{N} \delta_{(\Psi(\widehat{\mathbf{x}}_{(n)},\widehat{\mathbf{y}}_{(n)}))}$ where $\delta_{(\mathbf{x}^{\prime},\mathbf{y}^{\prime})}$ is the Dirac point mass located at $(\mathbf{x}^{\prime},\mathbf{y}^{\prime})$, and equipped with transportation cost
\begin{equation*}
    d_{\mathbb{H}}(\boldsymbol{\xi}_{(\mathbb{H})},\boldsymbol{\xi}_{(\mathbb{H})}^{\prime}) \coloneqq || \mathbf{x}_{(\mathbb{H})} - \mathbf{x}_{(\mathbb{H})}^{\prime} ||_{\mathbb{H}} + \kappa \mathnormal{1}_{\{\mathbf{y} \neq \mathbf{y}^{\prime} \}}.
\end{equation*}

Next, let us consider the function $g(\mathbf{x}_{(\mathbb{H})}) \coloneqq \mathbf{g}_{(\mathbb{H})_i} - \mathbf{g}_{(\mathbb{H})_j}$ for any $i,j \in [C]$. We can obtain the Lipschitz modulus of this function as follows:
    \begin{subequations}
    \begin{align}
        \Lip(g(\mathbf{x}_{(\mathbb{H})})) &\label{eq:t2_1_0}= \sup_{\boldsymbol{\omega}} \left \{ || \boldsymbol{\omega} ||_{\ast} \colon g^{\ast}(\mathbf{x}_{(\mathbb{H})}) < \infty \right \}\\
        &\label{eq:t2_1_1}= \sup_{\boldsymbol{\omega}} \left \{ || \boldsymbol{\omega} ||_{\ast} \colon \sup_{\mathbf{x}_{(\mathbb{H)}}} \left \{ \langle \boldsymbol{\omega}, \mathbf{x}_{(\mathbb{H})} \rangle_{\mathbb{H}} -  \langle \mathbf{h}_{(i)}, \mathbf{x}_{(\mathbb{H})} \rangle_{\mathbb{H}} + \langle \mathbf{h}_{(j)}, \mathbf{x}_{(\mathbb{H})} \rangle_{\mathbb{H}} \right \} < \infty \right \}\\
        &\label{eq:t2_1_2}= \sup_{\boldsymbol{\omega}} \left \{ || \boldsymbol{\omega} ||_{\ast} \colon \sup_{\mathbf{x}_{(\mathbb{H)}}} \left \{ \langle \boldsymbol{\omega}, \mathbf{x}_{(\mathbb{H})} \rangle_{\mathbb{H}} - \langle \mathbf{h}_{(i)} - \mathbf{h}_{(j)}, \mathbf{x}_{(\mathbb{H})} \rangle_{\mathbb{H}} \right \} < \infty \right \}\\
        &\label{eq:t2_1_3}= \sup_{\boldsymbol{\omega}} \left \{ || \boldsymbol{\omega} ||_{\ast} \colon \boldsymbol{\omega} = \mathbf{h}_{(i)} - \mathbf{h}_{(j)} \right \}\\
        &\label{eq:t2_1_4}= ||\mathbf{h}_{(i)} - \mathbf{h}_{(j)}||_{\mathbb{H}},
    \end{align}
    \end{subequations}
    where \eqref{eq:t2_1_0} follows from the definition of the Lipschitz modulus, \eqref{eq:t2_1_1} follows from the definition of the conjugate function, \eqref{eq:t2_1_3} follows by observing that this is the only condition under which the inner maximization problem is bounded, and finally \eqref{eq:t2_1_4} follows by recalling the fact that $||\cdot||_{\mathbb{H}}$ is self-dual. We note that this result is very similar to the one proved in Lemma \ref{lem:prop_loss} Part \ref{part:lip_mod}. Thus, we utilize this result in a similar fashion to derive the tractable upper bound of the kernelized version of the WDR-MSVM.

    We use the previous results to rewrite the lifted learning problem \eqref{eq:lifted_nonlin} as follows.
    \begin{subequations}
    \begin{align}
        &\inf_{\{ \mathbf{h}_{(c)}\}_{c=1}^C \subset \mathbb{H}} \sup_{\mathbb{Q} \in \mathcal{A}_{\varepsilon,1}^{\mathbb{H}}(\Xi)} \mathbb{E}^{\mathbb{Q}} [\ell_{\text{CS},\mathbb{H}}(\mathbf{g}_{(\mathbb{H})}(\mathbf{x}_{(\mathbb{H})});\mathbf{y})] \nonumber\\
        &\label{eq:t2_2_1} = \left \{ \begin{aligned}
            & \min_{\{\mathbf{h}_{(c)}\}_{c=1}^C \subset \mathbb{H},\lambda,s_n} && \lambda \varepsilon + \frac{1}{N} \sum_{n=1}^{N} s_n&& \\
            & \subjectto && \ell_{\text{CS},\mathbb{H}}(\mathbf{H}(\widehat{\mathbf{x}}_{(n)});\widehat{\mathbf{y}}_{(n)}) \leq s_n && \forall n \in [N] \\
            &&&  \ell_{\text{CS},\mathbb{H}}(\mathbf{H}(\widehat{\mathbf{x}}_{(n)});\widecheck{\mathbf{y}}_{(c)}) - \lambda \kappa \leq s_n && \forall n \in [N] \  \forall \widecheck{\mathbf{y}}_{(c)} \in \mathcal{Y}, \ \widecheck{\mathbf{y}}_{(c)} \neq \widehat{\mathbf{y}}_{(n)} \\
            &&& \lambda \geq \left | \left | (\mathbf{v}_{(a)} - \mathbf{v}_{(b)})^{\mathsf{T}}\mathbf{H} \right | \right |_2 && \forall b,c \in [C] \\
        \end{aligned} \right.\\
        & \label{eq:near_final_kernel} \leq \left \{ \begin{aligned}
            & \min_{\{\mathbf{h}_{(c)}\}_{c=1}^C \subset \mathbb{H},\lambda,s_n} && \lambda \varepsilon + \frac{1}{N} \sum_{n=1}^{N} s_n&& \\
            & \subjectto && \ell_{\text{CS},\mathbb{H}}(\mathbf{H}(\widehat{\mathbf{x}}_{(n)});\widehat{\mathbf{y}}_{(n)}) \leq s_n && \forall n \in [N] \\
            &&&  \ell_{\text{CS},\mathbb{H}}(\mathbf{H}(\widehat{\mathbf{x}}_{(n)});\widecheck{\mathbf{y}}_{(c)}) - \lambda \kappa \leq s_n && \forall n \in [N] \  \forall \widecheck{\mathbf{y}}_{(c)} \in \mathcal{Y}, \ \widecheck{\mathbf{y}}_{(c)} \neq \widehat{\mathbf{y}}_{(n)} \\
            &&& \lambda \geq \left | \left | \mathbf{h}_{(a)} \right | \right |_2 + \left | \left | \mathbf{h}_{(b)} \right | \right |_2 && \forall a,b \in [C]
        \end{aligned}\right.,
    \end{align}
    \end{subequations}
    where \eqref{eq:t2_2_1} follows by recalling the fact that $\mathbf{h}_{(c)_{(\mathbb{H})}}(\mathbf{x}_{(\mathbb{H})}) = \langle \mathbf{h}_{(c)},\mathbf{x}_{(\mathbb{H})} \rangle_{\mathbb{H}} =  \langle \mathbf{h}_{(c)},\Psi(\mathbf{x}) \rangle_{\mathbb{H}} = \mathbf{h}_{(c)}(\mathbf{x})$, and \eqref{eq:near_final_kernel} from the triangle inequality. Now, let us introduce the variable $\widehat{J}_{\mathbb{H}}$ that is equivalent to the optimal value of the problem in \eqref{eq:near_final_kernel}. Observe that $\widehat{J}_{\mathbb{H}}$ is non-decreasing in $||\mathbf{h}_{(i)}||_2$ for all $i \in [C]$. Moreover, the program in \eqref{eq:near_final_kernel} is a minimization problem, in which some of the decision variables are $\{ \mathbf{h}_{(i)}\}_{i=1}^C$. Thus, the Representer Theorem \cite{rep_thm} applies to our problem. This indicates that without sacrificing optimality, the optimal hypotheses can be written as $\mathbf{h}_{(i)}^{\ast}(\mathbf{x}) = \sum_{n=1}^{N}\alpha_nk(\mathbf{x}_{(n)},\mathbf{x})$. Finally, note that $\mathbf{h}_{(i)}(\widehat{\mathbf{x}}_{(n)}) = \sum_{j=1}^{N}\mathbf{A}_{ij}\mathbf{K}_{nj}$, and $||\mathbf{h}_{(i)}||_2^2 = \langle \mathbf{A}_{i \mathbf{\cdot}},\mathbf{K}\mathbf{A}_{i \mathbf{\cdot}} \rangle$ \cite{2019regularization}. Thus, the final claim follows by including $\mathbf{A}$ as a decision variable.

\end{proof}
\subsection{Proof of Theorem \ref{thm:alg}}
\begin{proof}
Firstly, observe that the Linear WDR-MSVM problem in Thm. \ref{thm:linear_DRMSVM} equipped with the $\ell_{\infty}$-norm can be rewritten as follows.

\begin{subequations}
\begin{align}
&\nonumber \inf_{\mathbf{M}} \sup_{\mathbb{Q} \in \mathcal{A}_{\varepsilon,1}(\Xi)} \mathbb{E}^{\mathbb{Q}} \left [ \ell_{\text{CS}}(\mathbf{M};\boldsymbol{\xi}) \right ] \\
    &=\left \{ \begin{aligned}
    &\min_{\mathbf{M},\lambda} && \lambda \varepsilon + \frac{1}{N} \sum_{n=1}^{N} \max \left \{ \ell_{CS} (\mathbf{M};(\widehat{\mathbf{x}}_{(n)},\widehat{\mathbf{y}}_{(n)})),\ell_{CS} (\mathbf{M};(\widehat{\mathbf{x}}_{(n)},\widecheck{\mathbf{y}}_{(0)})) - \lambda \kappa ,\dots,\ell_{CS} (\mathbf{M};(\widehat{\mathbf{x}}_{(n)},\widecheck{\mathbf{y}}_{(C)})) - \lambda \kappa\right \} \\ 
    & \subjectto && \lambda \geq \left | \left | (\mathbf{v}_{(i)} - \mathbf{v}_{(j)})^{\mathsf{T}}\mathbf{M} \right | \right |_{\infty} \qquad \forall i,j \in [C] 
    \end{aligned} \right.\\
    &\label{eq:alg1a}= \left \{ \begin{aligned}
        &\min_{\mathbf{M},\lambda}  
 &&\lambda \varepsilon + \frac{1}{N} \sum_{n=1}^{N} \max_{\mathbf{v}_{(c)} \in \mathcal{Y},\mathbf{y}_{(c)} \in \mathcal{Y}} \left \{ \left (\mathbf{v}_{(c)} \right )^\mathsf{T} \left ( \mathbf{M} \widehat{\mathbf{x}}_n - \mathbf{y}_{(c)} \right ) - \mathbf{y}_{(c)}^{\mathsf{T}} \mathbf{M} \widehat{\mathbf{x}}_n - \lambda \kappa \mathbf{1}_{\{\mathbf{y}_{(c)} \neq \widehat{\mathbf{y}}_n\}} \right \} \\
        & \subjectto && \lambda \geq \left | \left | (\mathbf{v}_{(i)} - \mathbf{v}_{(j)})^{\mathsf{T}}\mathbf{M} \right | \right |_{\infty} \qquad \forall i,j \in [C] 
    \end{aligned}\right.
\end{align}
\end{subequations}

Now, let
\begin{equation}\label{eq:obj_func_temp}
    f(\lambda,\mathbf{M}) \coloneqq \lambda \varepsilon + \frac{1}{N} \sum_{n=1}^{N} \max_{\mathbf{v}_{(c)} \in \mathcal{Y},\mathbf{y}_{(c)} \in \mathcal{Y}} \left \{ \left (\mathbf{v}_{(c)} \right )^\mathsf{T} \left ( \mathbf{M} \widehat{\mathbf{x}}_n - \mathbf{y}_{(c)} \right ) - \mathbf{y}_{(c)}^{\mathsf{T}} \mathbf{M} \widehat{\mathbf{x}}_n - \lambda \kappa \mathbf{1}_{\{\mathbf{y}_{(c)} \neq \widehat{\mathbf{y}}_n\}} \right \}.
\end{equation}
Observe that the convexity and Lipschitz continuity of $f(\lambda,\mathbf{M})$ follow from the CS loss properties proven in Lemma \ref{lem:prop_loss}. Thus, the projected subgradient method equipped with an appropriately diminishing stepsize can be utilized to solve the problem in \eqref{eq:alg1a} \cite{nesterov2013}. Moreover, one can directly see that
\begin{equation*}
    \varepsilon + \kappa \sum_{n=1}^{N} \mathbf{1}_{\{\tau(\lambda,\mathbf{M})\}}(n) \in \partial_{\lambda} f(\lambda,\mathbf{M}),
\end{equation*}
where $\mathbf{1}_{\{\tau(\lambda,\mathbf{M})\}}(n)$ is equivalent to $1$ if $\ell_{\text{CS}} (\mathbf{M};(\widehat{\mathbf{x}}_{(n)},\widecheck{\mathbf{y}}_{(c)})) - \lambda \kappa > \ell_{\text{CS}} (\mathbf{M};(\widehat{\mathbf{x}}_{(n)},\widehat{\mathbf{y}}_{(n)}))$ for any $\widecheck{\mathbf{y}}_{(c)} \in \mathcal{Y}, \widecheck{\mathbf{y}}_{(c)} \neq \widehat{\mathbf{y}}_{(n)}$, and is equivalent to $0$ otherwise. Similarly, one can directly observe that 
\begin{equation*}
    \sum_{n=1}^{N} \left (\mathbf{v}_{(c)}^{\ast}(n) - \mathbf{y}_{(c)}^{\ast}(n) \right )^{\mathsf{T}} \widehat{\mathbf{x}}_{(n)} \in \partial_{\mathbf{M}} f(\lambda,\mathbf{M}),
\end{equation*}
where $\mathbf{v}_{(c)}^{\ast}(n),\mathbf{y}_{(c)}^{\ast}(n) \coloneqq \argmax_{\mathbf{v}_{(c)}\in \mathcal{Y},\mathbf{y}_{(c)} \in \mathcal{Y}}  \left (\mathbf{v}_{(c)} \right )^\mathsf{T} \left ( \mathbf{M} \widehat{\mathbf{x}}_n - \mathbf{y}_{(c)} \right ) - \mathbf{y}_{(c)}^{\mathsf{T}} \mathbf{M} \widehat{\mathbf{x}}_n - \lambda \kappa \mathbf{1}_{\{\mathbf{y}_{(c)} \neq \widehat{\mathbf{y}}_n\}}$.

Therefore, the previous subgradients of $f(\lambda,\mathbf{M})$ in terms of $\lambda$ and $\mathbf{M}$ can be used to take subgradient steps on both variables. However, it is also required during each iteration to solve the projection problem $\Pi(\lambda^{\prime},\mathbf{M}^{\prime})$ defined as
\begin{subequations}
\begin{align}
\Pi(\lambda^{\prime},\mathbf{M}^{\prime}) &=  \left \{\begin{aligned}
    &\min_{\lambda,\mathbf{M}} &&(\lambda - \lambda^{\prime})^2 + \sum_{c=1}^{C} \sum_{p = 1}^{P} (\mathbf{M}_{cp} - \mathbf{M}_{cp}^{\prime})^2&&\\
    &\subjectto && \lambda \geq \left | \left | (\mathbf{v}_{(i)} - \mathbf{v}_{(j)})^{\mathsf{T}}\mathbf{M} \right | \right |_{\infty} && \forall i,j \in [C] 
\end{aligned} \right..\\
& = \left \{\begin{aligned}
    &\min_{\lambda,\mathbf{M}} &&(\lambda - \lambda^{\prime})^2 + \sum_{c=1}^{C} \sum_{p = 1}^{P} (\mathbf{M}_{cp} - \mathbf{M}_{cp}^{\prime})^2&&\\
    &\subjectto && \lambda \geq |(\mathbf{v}_{(i)} - \mathbf{v}_{(j)})^{\mathsf{T}}\mathbf{M}_{\cdot d}| && \forall i,j \in [C] , \forall d \in [P]
\end{aligned} \right..\\
& = \left \{\begin{aligned}
    &\min_{\lambda,\mathbf{M}} &&(\lambda - \lambda^{\prime})^2 + \sum_{c=1}^{C} \sum_{p = 1}^{P} (\mathbf{M}_{cp} - \mathbf{M}_{cp}^{\prime})^2&&\\
    &\subjectto && \lambda \geq |\mathbf{M}_{id} - \mathbf{M}_{jd}| && \forall i,j \in [C] , \forall d \in [P]
\end{aligned} \right..\\
&\label{eq:proj_fin} = \left \{\begin{aligned}
    &\min_{\lambda,\mathbf{M},m_d} &&(\lambda - \lambda^{\prime})^2 + \sum_{c=1}^{C} \sum_{p = 1}^{P} (\mathbf{M}_{cp} - \mathbf{M}_{cp}^{\prime})^2&&\\
    &\subjectto && m_d - \frac{\lambda}{2} \leq \mathbf{M}_{id} \leq m_d + \frac{\lambda}{2} && \forall i \in [C],\forall d \in [P]
\end{aligned} \right..\\,
\end{align}
\end{subequations}
where the problem in \eqref{eq:proj_fin} is obtained by realizing that all the elements within a column $d$ of a feasible $\mathbf{M}$ must be within $\lambda$ of each other. Therefore, they must be within $\lambda/2$ of the mean $m_d$ of the max and min entries of the column $d$. Since $m_d$ is not known, however, they are included as decision variables in the problem, leading to the final algorithm presented in the theorem.
\end{proof}
\subsection{Proof of Proposition \ref{prop:conv}}
\begin{proof}
Firstly, observe that the simplified projection problem $\Pi(\lambda^{\prime},\mathbf{M}^{\prime})$ show in Thm. \ref{thm:alg} can be written as a QCQP with a linear objective as follows:
\begin{equation}\label{eq:qcqp_proj}
\Pi(\lambda^{\prime},\mathbf{M}^{\prime}) = \left \{ \begin{aligned}
    &\min_{\lambda,\mathbf{M},m_d,s_{\lambda},\mathbf{S}} &&s_{\lambda} + \sum_{c=1}^{C} \sum_{p=1}^{P} \mathbf{S}_{cp}&&\\
    &\subjectto && s_{\lambda} \geq (\lambda - \lambda^{\prime})^2&&\\
    &&& \mathbf{S}_{cp} \geq (\mathbf{M}_{cp} - \mathbf{M}_{cp}^{\prime})^2 && \forall c \in [C], \forall p \in [P]\\
    &&&  m_d - \frac{\lambda}{2} \leq \mathbf{M}_{id} \leq m_d + \frac{\lambda}{2} && \forall i \in [C],\forall d \in [P]
\end{aligned}\right..
\end{equation}

Now, note that the problem in \eqref{eq:qcqp_proj} has $U = 3PC + 1$ constraints and $V = 2PC + P + 2$ decision variables. Therefore, the theoretical worst-case time complexity of solving the problem via the barrier method equipped with the log barrier and Newton updates would be $\mathcal{O}(P^{3.5}C^{3.5} \log(\beta \epsilon_2^{-1}))$, where $\epsilon_2$ is the optimality tolerance of the solution and $\beta$ is a data-dependent constant \cite{nn1994}. Moreover, note that computing the subgradients of $f(\lambda,\mathbf{M})$ defined in \eqref{eq:obj_func_temp} with respect to $\lambda$ and $\mathbf{M}$ requires $NC^2$ iterations to iterate over all the samples and possible combinations of $\mathbf{v}_{(c)}$ and $\mathbf{y}_{(c)}$. Each such iteration involves multiple arithmetic vector operations, the time complexity of each of which is either $P$ or $C$. Thus, the final result follows by noting that the projected subgradient method converges to a solution with optimality tolerance $\epsilon_1$ in $\mathcal{O}(\epsilon_1^{-2})$ iterations \cite{bubeck2015} assuming the stepsize $\sigma(t) \rightarrow 0$ as $t \rightarrow \infty$ and $\sum_{t=1}^{\infty} \sigma(t) = \infty$ \cite{nesterov2013}.
\end{proof}
\newpage
\section{Appendix B: Additional Time Complexity Details}
Equipped with the log barrier function and utilizing Newton updates, the barrier method can reach a solution within an optimality tolerance of $\epsilon$ in $\mathcal{O}(1)\sqrt{U}\log{(\beta\epsilon^{-1})}$, where $U$ is the number of constraints and $\beta$ is a data-dependent constant \cite{nn1994}. Moreover, the arithmetic costs of a Newton update for an LP and a QCQP are $\mathcal{O}(UV^2)$ and $\mathcal{O}([U+V]V^2)$, respectively, where $V$ is the number of decision variables \cite{nn1994}. We summarize the number of constraints and decision variables for the linear and kernel versions of our proposed WDR-MSVM model in the Tab. \ref{tb:d_and_c} as follows.

\begin{table}[H]
\centering
\begin{tabular}{llll} 
\hline
\textbf{Model}&\textbf{LP/QCQP}&\textbf{U}& \textbf{V}       \\ 
\hline
Linear, $\ell_{\infty}$-norm&LP&$(N+2P)C^2$& $CP+N+1$  \\ 
\hline
Linear, $\ell_{1}$-norm&LP&$(N+2P+1)C^2$& $CP+N+1$  \\ 
\hline
Linear, $\ell_{2}$-norm&QCQP&$(N+1)C^2$& $CP+N+1$  \\ 
\hline
Kernel&QCQP&$(N+1)C^2$& $CN+N+1$  \\
\hline
\end{tabular}
\caption{Number of constraints $U$ and decision variables $V$ for all different versions of our models proposed in Thms. \ref{thm:linear_DRMSVM} and \ref{thm:kernel_DRMSVM}.}
\label{tb:d_and_c}
\end{table}
\newpage
\section{Appendix C: Additional Numerical Experiments and Experimental Results}
\subsection{Detailed Sensitivity Analysis Results from Simulation Experiment 1}

\begin{figure}[H]
    \centering
    \includegraphics[width=0.92\textwidth]{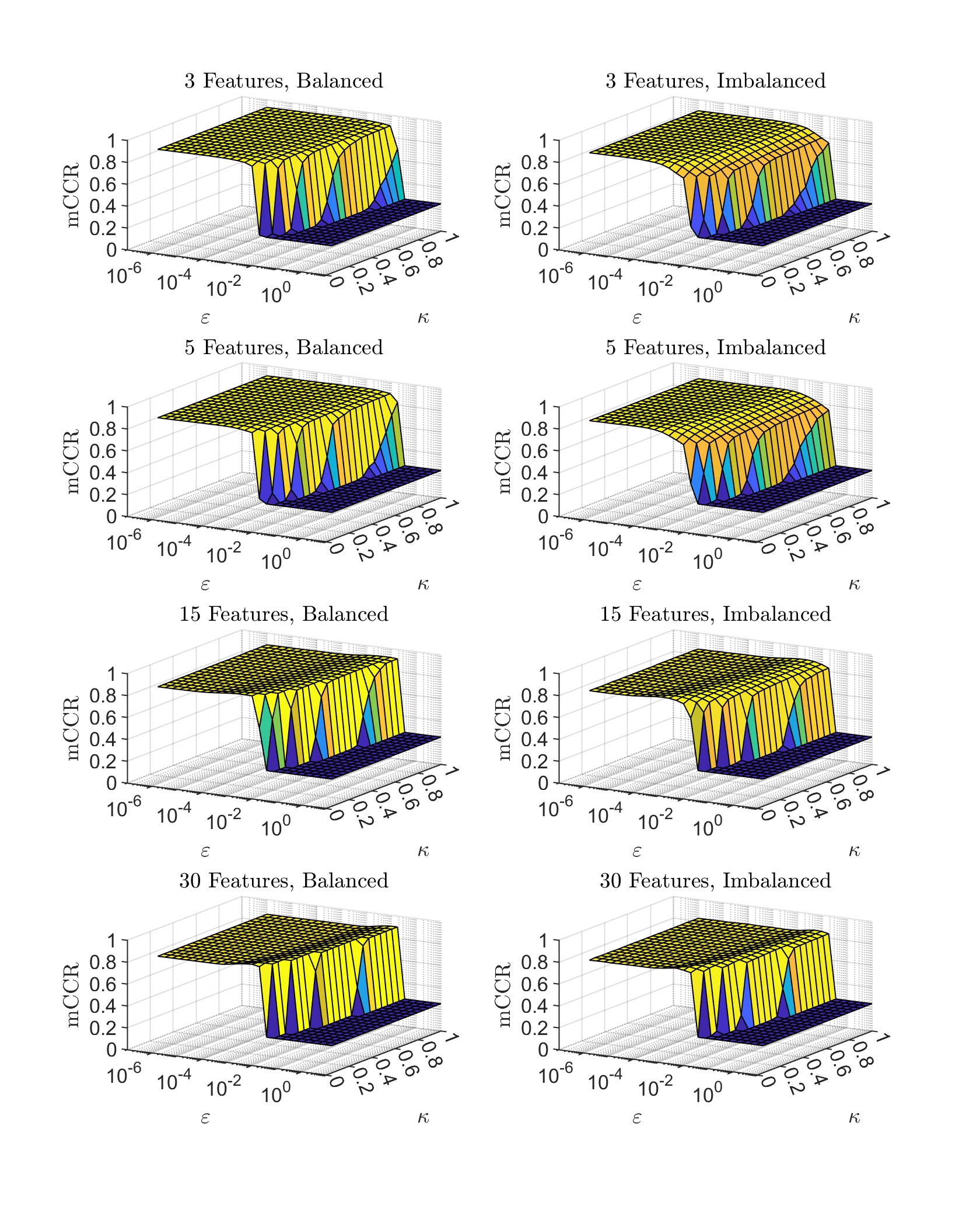}
    \caption{Surface plots of mCCR vs. $\varepsilon$ and $\kappa$ for the linear WDR-MSVM with 4 classes.}
    \label{fig:sur_msvm4}
\end{figure}

\begin{figure}[H]
    \centering
    \includegraphics[width=0.93\textwidth]{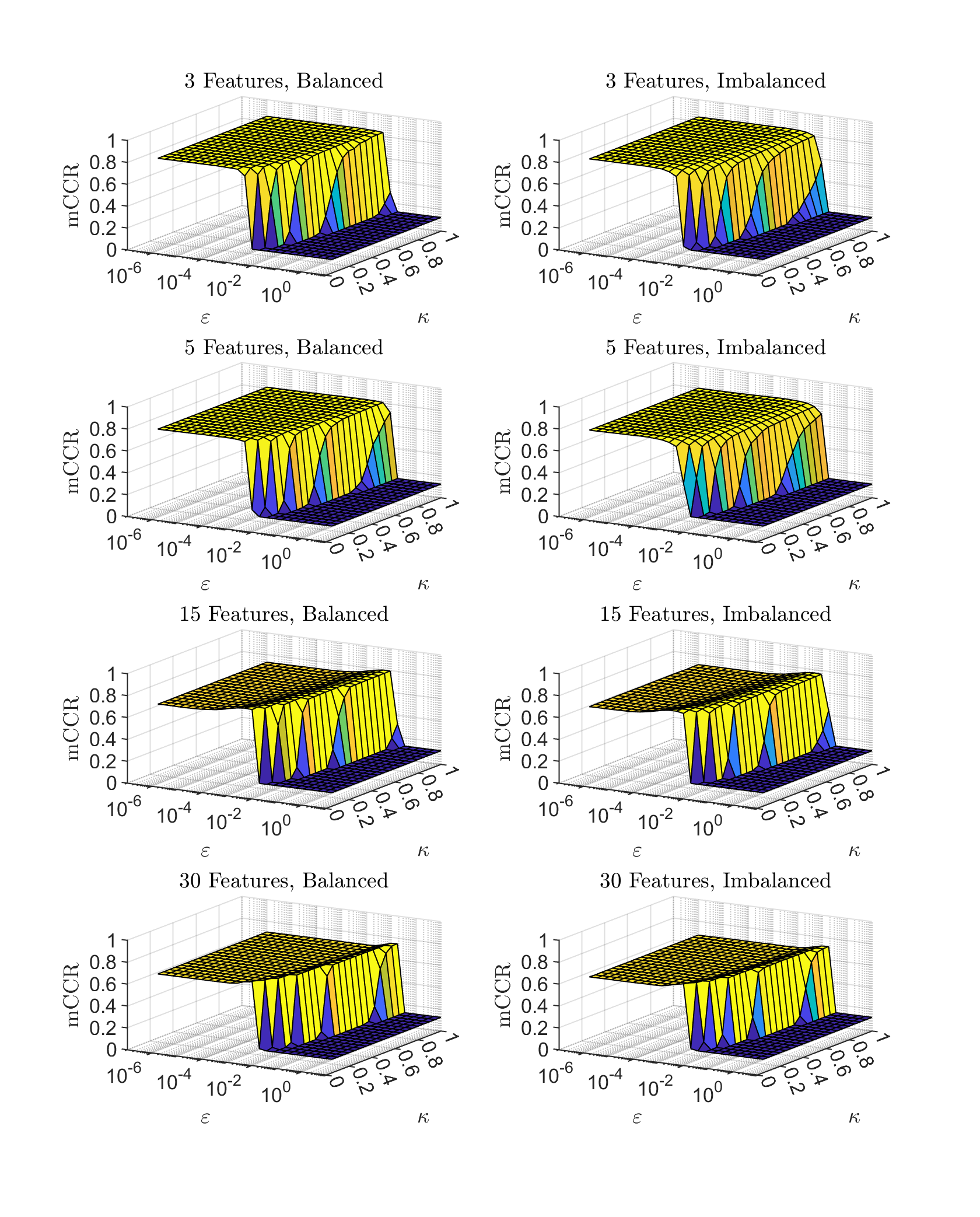}
    \caption{Surface plots of mCCR vs. $\varepsilon$ and $\kappa$ for the linear WDR-MSVM with 8 classes.}
    \label{fig:sur_msvm8}
\end{figure}

\begin{figure}[H]
    \centering
    \includegraphics[width=0.93\textwidth]{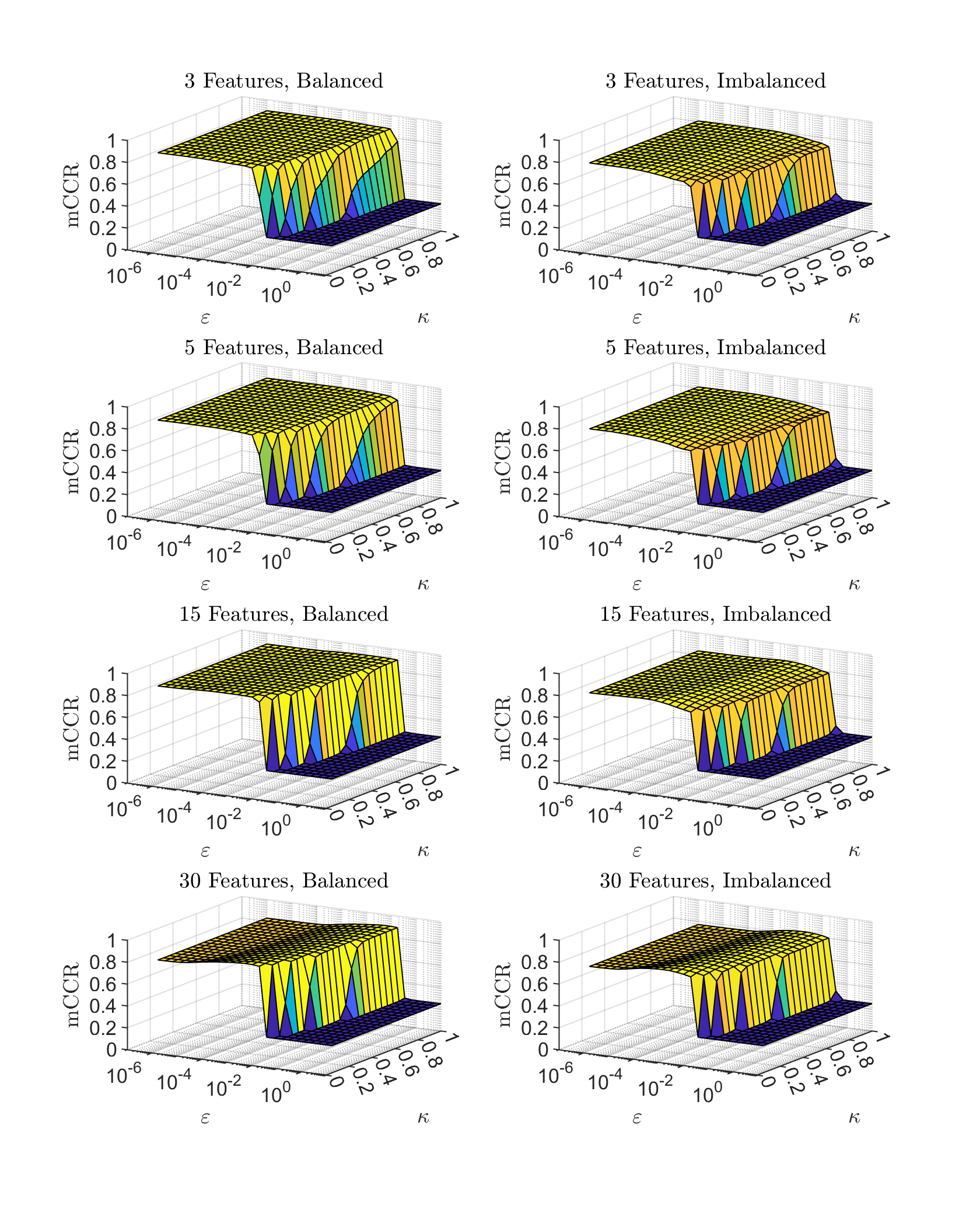}
    \caption{Surface plots of mCCR vs. $\varepsilon$ and $\kappa$ for the linear DR-OVA with 4 classes.}
    \label{fig:sur_ova4}
\end{figure}

\begin{figure}[H]
    \centering
    \includegraphics[width=0.93\textwidth]{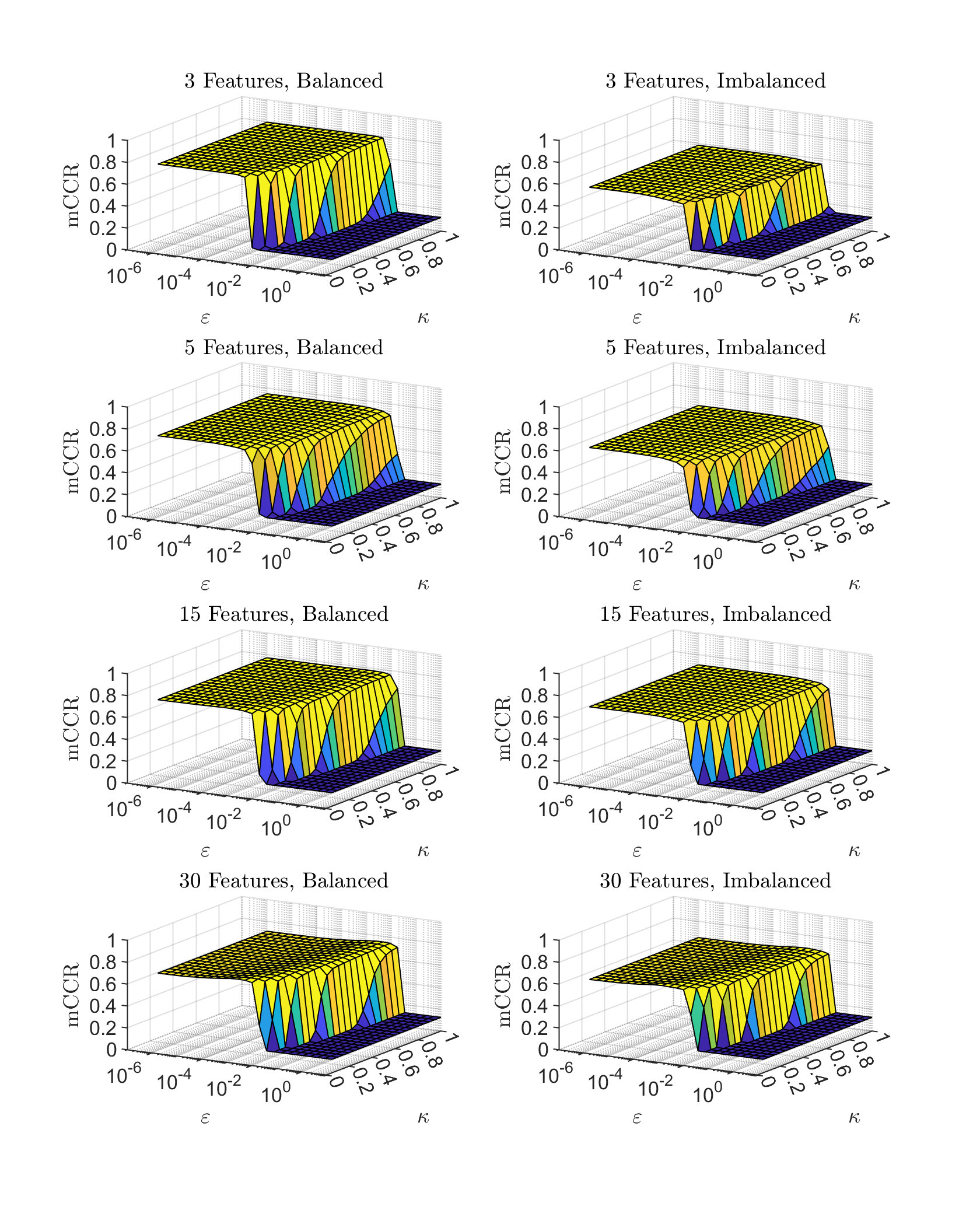}
    \caption{Surface plots of mCCR vs. $\varepsilon$ and $\kappa$ for the linear DR-OVA with 8 classes.}
    \label{fig:sur_ova8}
\end{figure}

\subsection{Experiment 3: Scalability Experiment}
 We executed an empirical study to evaluate the running times in solving both the linear and kernel versions of our model and their regularized counterparts as the size of the training dataset grows. More specifically, we independently examined three experimental settings:
\begin{itemize}
\item Increasing number of classes: $N = 360$, $P = 10$, $C \in \{4,6,8,10\}$.
\item Increasing number of features: $N = 360$, $C = 8$, $P \in \{4,10,16,22,28\}$.
\item Increasing number of training samples: $P = 10$, $C = 4$, $N \in \{80,360,640,920,1200\}$.
\end{itemize}
We implemented and solved the linear model equipped with the $\ell_{\infty}$-norm and the kernel model equipped with the radial basis function (RBF) kernel using the barrier method in Gurobi. We used solution tolerance of $\epsilon = 1\times10^{-2}$ and recorded the runtime to solve the problem for each run from the above experimental combinations. We repeated each combination 50 times with data randomly generated using the \texttt{make\_classification} module from the scikit-learn Python package \cite{scikit-learn}, and computed the mean runtime for each combination.

\begin{figure*}[t]
\centering
\begin{subfigure}{0.99\textwidth}
    \includegraphics[width=\textwidth]{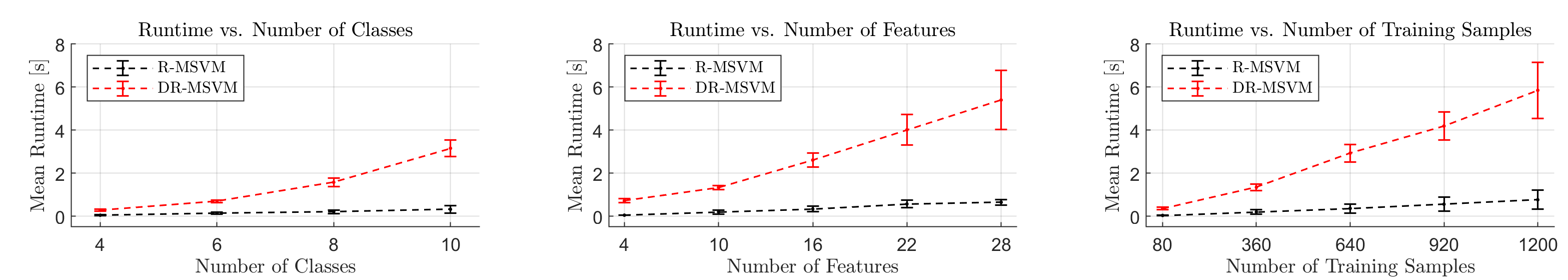}
    \caption{Plots of runtime vs. number of classes $C$, number of features $P$, and number of training samples $N$ for the simulation experiment with linear models.}
    \label{fig:scal_lin}
\end{subfigure}
\hfill
\begin{subfigure}{0.99\textwidth}
    \includegraphics[width=\textwidth]{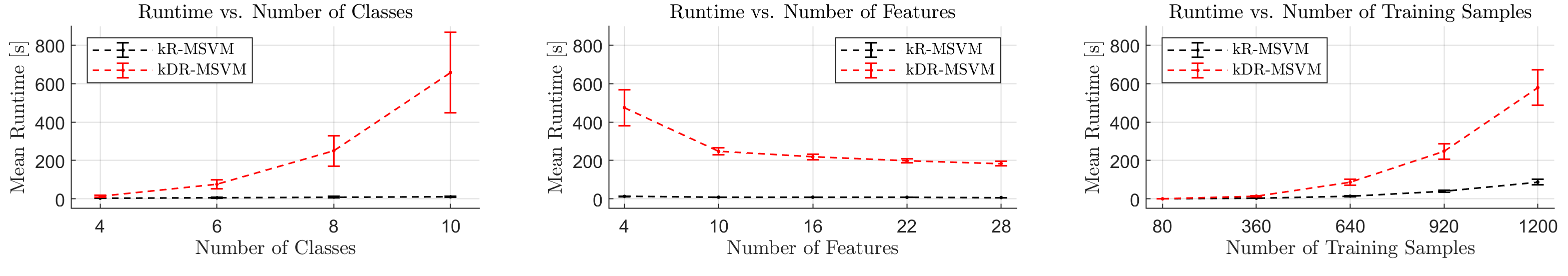}
    \caption{Plots of runtime vs. number of classes $C$, number of features $P$, and number of training samples $N$ for the simulation experiment with RBF kernel models.}
    \label{fig:scal_ker}
\end{subfigure}
\caption{Results of the scalability experiments.}
\end{figure*}

As expected, the linear and kernel versions of our DR model exhibit higher runtime than their regularized counterparts due to their more complex structure. Thus, the choice between the DR and the regularized versions of the model is a trade-off between model accuracy and training time. Indeed, the DR model would be more suited to applications that are highly sensitive to model accuracy and where ample time is available for training. Interestingly, we observe that for the kernel models, the mean runtime initially decreases and then remains constant as $P$ increases. Theoretically, the runtime should not depend on $P$ since this dimension is abstracted away from the training data when the kernel is applied. Thus, the change in runtime is solely dependent on the kernel parameter $\gamma$. Indeed an inappropriate value for $\gamma$ can lead to a poorly fit model, which may take a very long time to converge. In our study, we utilized the commonly-used $\gamma = 1/P$. However, it is possible that for the case where $P=4$ this value of $\gamma$ is not appropriate, resulting in the higher-than-expected runtime.

\subsection{Experiment 4: Projected Subgradient Method Algorithm Scalability}
In this experiment we seek to empirically evaluate if the proposed Alg. \ref{alg:subgrad} is indeed more scalable to large-scale problems than the use of the barrier method in a solver. However, due to limited computational resources, we do not perform experiments on large scale datasets. Instead, we assess scalability via small to medium-scale problems. This experiment uses data generated via the \texttt{make\_classification} module of the scikit-learn Python package \cite{scikit-learn}. The data generated is identical to that used in the previous scalability experiment. In this experiment we explore three settings:
\begin{itemize}
\item Increasing number of classes: $N_{Test} = 2000$, $N = 1000$, $P = 4$, $C \in \{4,6,8,10,12\}$.
\item Increasing number of features: $N_{Test} = 2000$, $N = 1000$, $C = 4$, $P \in \{4,10,16,22,28\}$.
\item Increasing number of training samples: $N_{Test} = 2000$, $P = 4$, $C = 4$, $N \in \{1000,2000,3000,4000,5000\}$.
\end{itemize}

For both solution algorithms, we utilize $\varepsilon = 1\times 10^{-4}$, $\kappa = 0.5$, and the $\ell_{\infty}$-norm. For the barrier method solution we use an optimality tolerance of $1\times10^{-2}$, whereas we use $T = 140$ iterations for the subgradient algorithm. Moreover, for the subgradient algorithm we evaluate accuracy in each experimental setting for initial stepsize $\sigma(0) \in \{1\times10^{-2},1\times10^{-1},1\times10^{0},1\times10^{1},1\times10^{2}\}$, and we set $\sigma(t) = \sigma(0)/(t)$. We only report results for the $\sigma(0)$ attaining the highest final model accuracy. Moreover, we initialize $\lambda^{(0)} = 0$ and $\mathbf{M}^{(0)} = \mathbf{0}$ for the subgradient algorithm. We repeat each experimental run $50$ times. In each experimental run we evaluate the ratio between the training runtimes of the barrier method and our proposed subgradient algorithm, as well as the CCR of the models attained by both methods. Note that we utilize a fixed number of iterations for our proposed subgradient algorithm instead of a stopping criterion as it is documented in the literature that the subgradient method does not have a practically implementable stopping criterion \cite{Bagirov2014}. However, our reporting of the mCCR attained by the model serves as a check that the final trained model is near-optimal, and effectively usable in practice. In a real-world setting, both the number of iterations $T$ and the initial stepsize $\sigma(0)$ can be treated as a model hyperparamters, and can be chosen via cross-validation.

\begin{figure*}[t]
\centering
\begin{subfigure}{0.99\textwidth}
    \includegraphics[width=\textwidth]{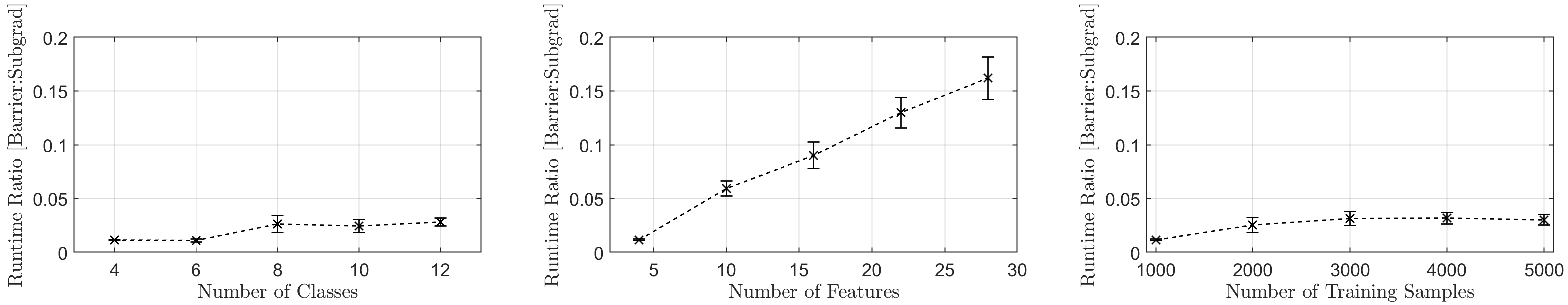}
    \caption{Plots of runtime ratio of both training algorithms (Barrier:Subgradient) vs. number of classes $C$, number of features $P$, and number of training samples $N$.}
    \label{fig:subgrad_ratio}
\end{subfigure}
\hfill
\begin{subfigure}{0.99\textwidth}
    \includegraphics[width=\textwidth]{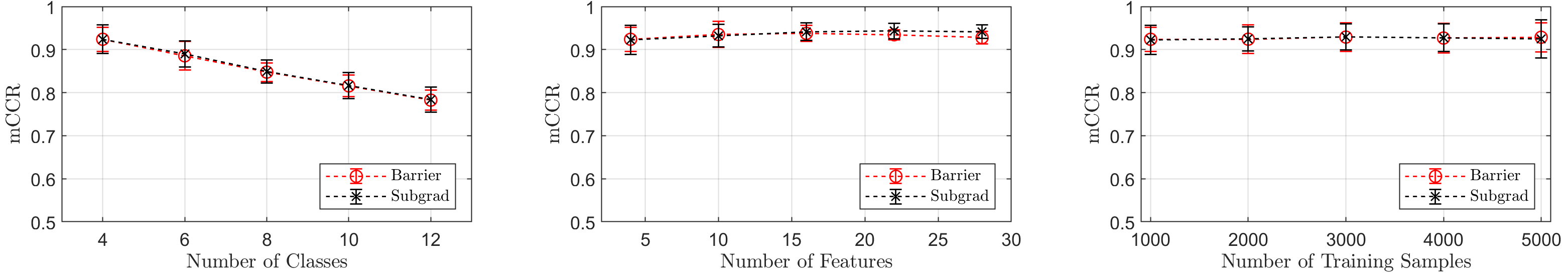}
    \caption{Plots of mCCR of both training algorithms vs. number of classes $C$, number of features $P$, and number of training samples $N$.}
    \label{fig:subgrad_acc}
\end{subfigure}
\caption{Results of the projected subgradient algorithm scalability experiments.}
\end{figure*}

Figures \ref{fig:subgrad_ratio} and \ref{fig:subgrad_acc} demonstrate the mean runtime ratio and mCCR over all $50$ experimental runs for all settings explored. We begin by studying Figure \ref{fig:subgrad_ratio}. We observe that the runtime ratio exhibits the most prominent increasing trend when the number of features $P$ increases. This prominent trend suggests that at a high enough $P$ the runtime ratio will exceed $1$, making our proposed algorithm more scalable to problems with a very large number of features. A less prominent trend of increasing runtime ratio is observed as the number of classes $C$ increases. While the existence of a trend suggests that at a high enough $C$ the ratio will exceed $1$, future research with more computational resources should focus on confirming this on datasets with a very large $C$ to confirm this hypothesis. Finally, no distinguishable trend could be observed in the runtime ratio as the number of training samples $N$ increases. This may be counter-intuitive as the theoretical worst-case time complexity computed for the projected subgradient algorithm in Prop. \ref{prop:conv} implies that it should be more scalable in $N$ than the barrier method. However, we note that worst-case time complexity estimates need not manifest in practice, specially for small to medium-scale problems. Therefore, future research should repeat this experiment with datasets with a very large $N$ to investigate if the theoretical improvement in time complexity attained by our proposed algorithm also holds in practice. We note, however, that even if our proposed algorithm does natively improve scalability with respect to $N$, it is still susceptible to stochastic subgradient approaches. Since such approaches utilize random batches of the data in each iteration, implementing them for datasets with a very large $N$ would result in an improvement in scalability.

Finally, we study Figure \ref{fig:subgrad_acc}. We observe that for all experimental settings, both training algorithms resut in models that attain almost identical mCCR values over the test set. This suggests that while a stopping criterion was not used for our proposed algorithm, $140$ iterations was indeed sufficient to converge to an optimal model.
\newpage
\section{Appendix D: Software, Hardware, and Dataset Details for Numerical Experiments}
\subsection{Hardware Details}
All the numerical experiments performed in this paper were run on Intel Xeon Gold 6226 CPUs @ 2.7 GHz (2 cores) with 10 Gb per core of DDR4-2933 MHz DRAM on a Linux operating system.
\subsection{Software Details}
We provide a list of all the software used in executing the numerical experiments in Tb. \ref{table:software}.
\begin{table}[H]
\centering
\begin{tabular}{lll}
\hline
\textbf{Software}                                 & \textbf{Version} & \textbf{License}            \\ \hline
Gurobi                                            & 10.0.1           & Academic Named-User License \\ \hline
Python                                            & 3.10.9           & PSF License                 \\ \hline
MATLAB                                            & 9.11             & Academic License            \\ \hline
\texttt{scikit-learn} Python Package  \cite{scikit-learn}                     & 1.5.1            & BSD License                 \\ \hline
\texttt{numpy} Python Package       \cite{numpy}                       & 1.23.5           & BSD License                 \\ \hline
\texttt{scipy} Python Package    \cite{scipy}                          & 1.10.0           & BSD License                 \\ \hline
\texttt{pandas} Python Package   \cite{pandas2}                          & 1.5.3            & BSD License                 \\ \hline
\texttt{ucimlrepo} Python Package                          & 0.0.3            & MIT License                 \\ \hline
Regularization via Mass Transportation Paper Code \\ \cite{2019regularization} & N/A              & MIT License                 \\ \hline
\end{tabular}
\caption{Details about software used in all numerical experiments.}
\label{table:software}
\end{table}

\subsection{Dataset Details}
We provide details on all the datasets used in our numerical experiments in Tb. \ref{table:datasets}.

\begin{table}[H]
\centering
\begin{tabular}{lll}
\hline
\textbf{Dataset} & \textbf{Notes}             & \textbf{License} \\ \hline
Wine             & N/A                        & CC By 4.0        \\ \hline
Penguins         & N/A                        & CC0 1.0          \\ \hline
Iris             & Nonlinear Class Separation & CC By 4.0        \\ \hline
Seeds            & N/A                        & CC By 4.0        \\ \hline
AI4I             & High Class Imbalance       & CC By 4.0        \\ \hline
\end{tabular}
\caption{Details on all real-world datasets used in numerical experiments.}
\label{table:datasets}
\end{table}

\end{document}